\documentclass[lettersize,journal]{IEEEtran}
\usepackage{amsmath,amsfonts}
\usepackage{amsmath,amssymb}
\usepackage{multirow}
\usepackage{subfigure}
\usepackage[linesnumbered,ruled,vlined,onelanguage]{algorithm2e}
\usepackage{color}
\usepackage{array}
\usepackage[caption=false,font=normalsize,labelfont=sf,textfont=sf]{subfig}
\usepackage{textcomp}
\usepackage{stfloats}
\usepackage{url}
\usepackage{verbatim}
\usepackage{graphicx}
\usepackage{cite}
\begin{document}

\title{Real-Time Text Detection with Similar Mask in Traffic, 
	Industrial, and  Natural Scenes }

\author{
	Xu~Han,  
	Junyu~Gao,~\IEEEmembership{Member,~IEEE,}
	Chuang~Yang, 
	Yuan~Yuan,~\IEEEmembership{~Senior Member,~IEEE}
	
	and Qi~Wang,~\IEEEmembership{~Senior Member,~IEEE}
	\thanks{
		
		X. Han, C. Yang are with the School of Computer Science, and with the School of Artificial Intelligence, Optics and Electronics (iOPEN), Northwestern Polytechnical University, Xi'an {\rm 710072}, China (E-mail: hxu04100@gmail.com, omtcyang@gmail.com).

	}
	\thanks{J. Gao, Y. Yuan, and Q. Wang are with the School of Artificial Intelligence, Optics and Electronics (iOPEN), Northwestern Polytechnical University, Xi'an {\rm 710072},  China (E-mail: gjy3035@gmail.com, y.yuan1.ieee@gmail.com, crabwq@gmail.com).}
	\thanks{This work was supported by the National Natural Science Foundation
	of China under Grant U21B2041, 62471394, 62306241. Qi Wang is the
	corresponding author.}
	}
\markboth{{IEEE} TRANSACTIONS ON INTELLIGENT TRANSPORTATION SYSTEMS}%
{Shell \MakeLowercase{\textit{et al.}}: A Sample Article Using IEEEtran.cls for IEEE Journals}
\maketitle

\begin{abstract}
Texts on the intelligent transportation scene include mass information. Fully harnessing this information is one of the critical drivers for advancing intelligent transportation. Unlike the general scene, detecting text in transportation has extra demand, such as a fast inference speed, except for high accuracy. Most existing real-time text detection methods are based on the shrink mask, which loses some geometry semantic information and needs complex post-processing. In addition, the previous method usually focuses on correct output, which ignores feature correction and lacks guidance during the intermediate process. To this end, we propose an efficient multi-scene text detector that contains an effective text representation similar mask (SM) and a feature correction module (FCM). Unlike previous methods,  the former aims to preserve the geometric information of the instances as much as possible. Its post-progressing saves 50$\%$ of the time, accurately and efficiently reconstructing text contours. The latter encourages false positive features to move away from the positive feature center, optimizing the predictions from the feature level. Some ablation studies demonstrate the efficiency of the SM and the effectiveness of the FCM. Moreover, the deficiency of existing traffic datasets (such as the low-quality annotation or closed source data unavailability) motivated us to collect and annotate a traffic text dataset, which introduces motion blur. In addition, to validate the scene robustness of the SM-Net, we conduct experiments on traffic, industrial, and natural scene datasets. Extensive experiments verify it achieves (SOTA) performance on several benchmarks. The code and dataset are available at: \url{https://github.com/fengmulin/SMNet}.
\end{abstract}


\section{Introduction}
\IEEEPARstart{S}{cene} text detection is a foundational task in computer vision, focused on pinpointing text occurrences. While text detection and recognition in documents have advanced significantly, scene texts present challenges due to their diverse attributes like colors, sizes, fonts, shapes, orientations, and complex backgrounds. It is an essential prerequisite for miscellaneous applications, such as automatic driving,  scene analysis \cite{traffic}, and intelligent transportation, so it attracts great interest in intelligent transportation communities. 

\begin{figure}[!t]
	\centering
	\includegraphics[width=1.0\linewidth]{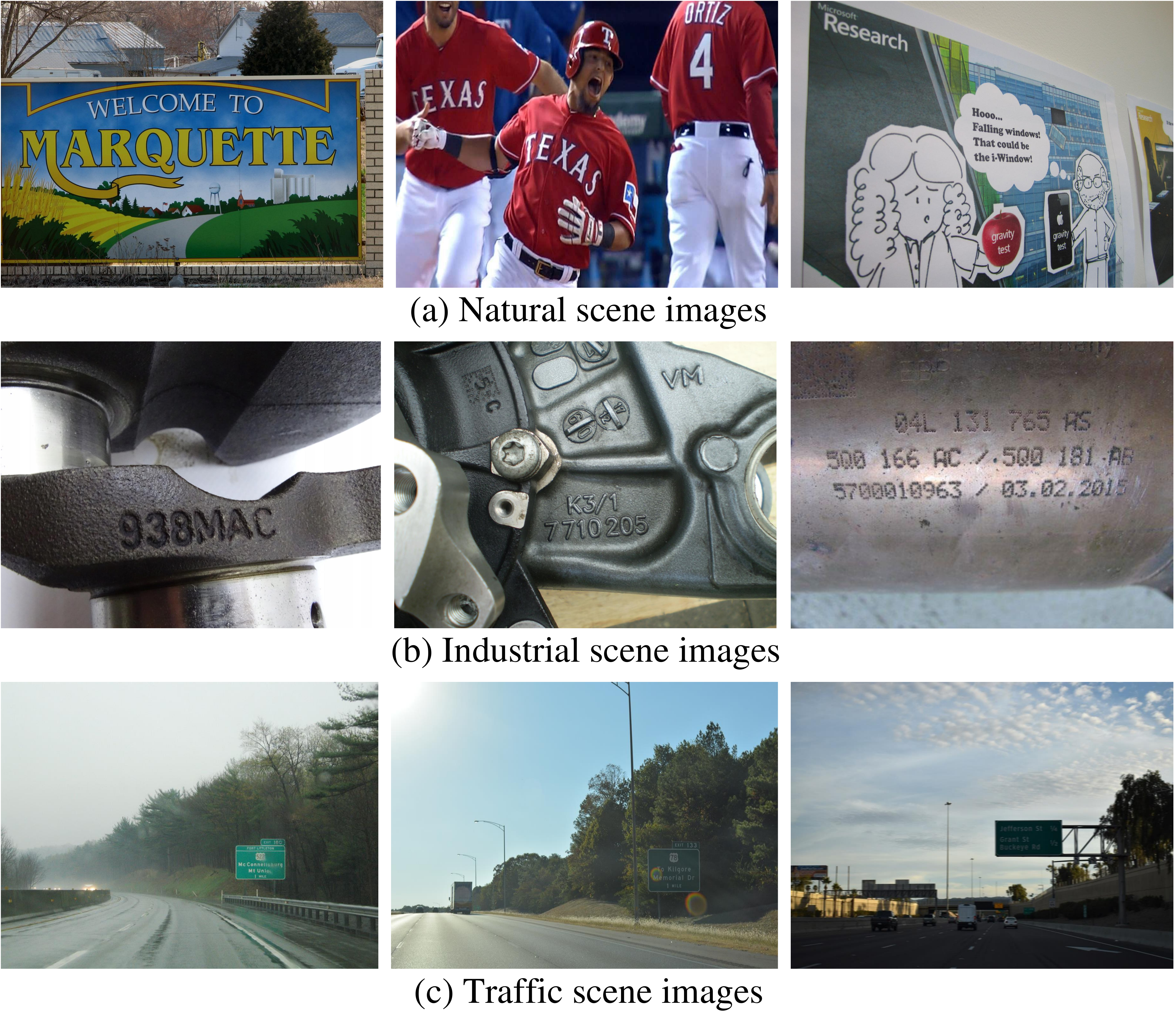}
	\caption{ Texts in different scenes. Texts in the natural scene enjoy a complex background. In the industrial scene,  low visual contrast and corroded surfaces make the difficult to detect text accurately.  Motion blur and changes in weather and lighting are the main challenges for traffic scenes.}
	\label{fig_1}
\end{figure}

Recently, scene text detection has made great strides due to the rapid advancements in semantic segmentation, instance segmentation, and object detection. However, numerous existing methods ignore some particular scenes, such as transportation and industrial manufacturing, warranting further exploration. Specifically, detecting text in industrial and traffic scenes proves more challenging than in natural scenes. Industrial scenes enjoy complex backgrounds, low visual contrast, and corroded surfaces. Weather, light, and motion blur influence traffic scene text detection more than other scenes. Some images from different scenes are visualized in Fig. \ref{fig_1}. Moreover, due to their unique characteristics, industrial and traffic scenes demand both high accuracy and rapid inference speed, compared to general scenes. The other challenge is that there are relatively few datasets and studies in these fields. For the industrial scene, a challenging industrial text detection dataset and corresponding synthetic datasets \cite{rfn} are proposed.  For the intelligent transportation scene,  there is a dataset \cite{tgpd} that only the test set is available, and a no-publicized transportation benchmark \cite{zhu2017cascaded}. To further promote the development of transportation text detection, we establish a new dataset that considers the specific motion blur on traffic scenes, which includes 1,000 training images and 528 testing images. The proposed motion blur traffic scene text (MBTST) dataset encompasses various challenges found in traffic scenes and simulates the real world as much as possible.

Apart from accuracy, high efficiency is crucial for transportation and industrial scene text detection, making segmentation-based methods the preferred option.  The majority of these methods rely on predicting shrink masks and extending them to reconstruct text instances.  The shrink mask is generated by shrinking the text region inward by a distance, which is computed by the area and perimeter of the instance.   Specificially, PSENet \cite{pse} predicts different scale shrink masks and utilizes a progressive scale expansion algorithm to recover text contour, which is accurate but inefficient. Based on it, PAN \cite{pan} proposes a pixel aggregation method to speed up the post-processing. DBNet \cite{db} introduces the Vatti clipping algorithm \cite{vatti1992generic} to further simplify the post-processing. ADNet \cite{ad} and RSMTD \cite{rmstd} utilize the prediction of the neural network to replace the computing of the expansion distance.
However, this shrinking method possesses two disadvantages: 1) It reconstructs text instances requiring complex operations like instance-level extendtion, area, and perimeter calculations. 2) It is an artificially defined concept with ambiguous definitions that humans cannot recognize its edge accurately, let alone models. Compared to the text region enjoys complete semantic features, it loses part geometry features associated with the instance contour. To address these limitations, we propose a new shrink calculation method coupled with efficient post-processing. Compared with the previous methods, it has several dominances:  1) Its simplistic post-progressing significantly improves overall efficiency. 2) Furthermore, it maximally preserves the geometric features of the text contour, which helps the model accurately recover them.

The previous methods solely guide and adjust the model based on the gap between output and ground truth, lacking sufficient guidance for the intermediate process. The effective transfer of knowledge between input and output proves challenging due to the numerous interspersed modules. This process is similar to the teacher explaining challenging questions to students. If only the answer is provided to students without guidance during the intermediate steps, it may reduce students' generalization ability. Specifically, when faced with a difficult problem, if the teacher only tells the correct result to the student with no intermediate steps, the student can only cope well with this kind of problem and not with similar ones. Inspired by this, to address this issue, we introduce a feature correction module that enhances the model at the feature level. This module encourages false positive features to keep away true positive ones, aiding the model in better distinguishing foreground and background, thereby significantly enhancing detection performance.

Based on the above method, an effective and efficient multiple-scene text detector, which achieves SOTA performance on multiple benchmarks, is proposed.
The main contributions of this paper are as follows:

\begin{enumerate}
	\item A new text representation called similar mask (SM) is proposed, which retains geometry shape information and demonstrates greater robustness compared to prior methods. Furthermore, it incorporates an extremely simplified post-processing step, reducing computation on the CPU, and enhancing detection efficiency.

	\item A feature correction module (FCM) is proposed, which guides the model in the intermediate process to suppress false positive predictions further. It helps the module distinguish foreground and background to improve detection performance without extra computation.

	\item  An efficient text detector  SM-Net is proposed, which achieves SOTA performance across multiple datasets encompassing transportation, industrial, and various scenes. Moreover, this method successfully strikes a trade-off between accuracy and efficiency, enabling its seamless application in real-world scenarios.
	
	\item We establish a transportation text dataset featuring diverse weather and illumination backgrounds. To further simulate real scenarios, we introduce motion blur to the images. In contrast to existing traffic text datasets, this incorporation of motion blur better mirrors the performance of various methods in real-world scenarios, significantly heightening the challenge.

\end{enumerate}

The remainder of this paper is structured as follows. Section \ref{relate} provides a review of related work on text detection in natural, industrial, and traffic scenes. In section \ref{dataser}, we describe the established traffic scene text dataset in detail. We illustrate the proposed method and the used loss function in section \ref{method}. A series of experiments are performed in section \ref{experiments} to demonstrate the superiority and effectiveness of the SM-Net. This paper is concluded in section \ref{conclusion}.
\section{Related Work}
\label{relate}
In this section, we overview the related work about text detection in natural, industrial, and traffic scenes.

\subsection{Text Detection in Natural Scene}
Existing deep learning text detection methods can be roughly divided into: regression-based, segmentation-based, and connect-component-based. 

Most previous regression-based text detection methods are based on general object detection frameworks like Faster-RCNN \cite{faster}. For example, based on it, Ma \emph{et al.} \cite{rrpn} proposed a modified pipeline RRPN, which further predicts the orientation of the text instance.  Zhou \emph{et al.} \cite{east} proposed EAST that treats the text contour as rotated rectangles or quadrangles to predict their bounding boxes. To cope with the varied aspect ratio of scene texts, Liao \emph{et al.} proposed TextBoxes \cite{textbox} based on SSD \cite{ssd}, which revises the convolution kernels and the default anchors. Furthermore, Liao \emph{et al.} proposed TextBoxes++ \cite{textbox++} that adds the angle parameters. Wang \emph{et al.} \cite{dmpn} first introduces prior quadrilateral sliding windows to detect multi-oriented instances, which significantly improves the detection performance. Liao \emph{et al.} \cite{rrd} proposed RRD that designed a hierarchical inception module and a text attention module to cope with the varied text scales and suppress background interference.  ABCNet \cite{abcnet} and FCENet \cite{fcenet} represented text instances by the Bezier curve and Fourier contour embedding, respectively. Although these methods achieve advanced performance, the detection speed is still a significant limit.

The key of the connect-component-based method is finding the text's parts or characters and concating them. Long \emph{et al.} \cite{textsnake} proposed TextSnake that represents text parts by a series of circles and joints them to recover instances. Similarly, Feng \emph{et al.} \cite{textdragon} switched to another scheme to utilize rectangles to represent text and dubbed it TextDragon. Zhang \emph{et al.} \cite{drrg} proposed DRRG, which introduces graph convolutional networks (GCN) to model the relationships between the predicted text components. Shi \emph{et al.} \cite{seglink} represented instances as segments and links, which adopt a prediction scheme to judge which segments belong to the same instances. The above methods need complex post-progressing, which limits the detection speed significantly and is not suitable for traffic and industrial scene text detection.

Inspired by FCN, many segmentation-based methods have emerged. Deng \emph{et al.} \cite{pixellink} predicted text region at pixel level and linked pixels belonging to the same instance. Then, they perform post-progressing to reconstruct instances. Wang \emph{et al.} \cite{pse} predicted different scale shrink masks and gradually expanded the shrink mask from minimal to text region.  STD \cite{spot} imitated spotlight to encourage the model to focus on candidate regions to suppress false positives. FEPE \cite{fepe} proposed a method to focus on region and instance features. LeafText \cite{leaftext} proposed a novel text representation method inspired by leaf vein biomimicry. TKC \cite{tkc} proposed a new text kernel calculation method based on the short side of the text. All three of these methods require complex post-processing. To speed up the inference, based on PSE-Net, Wang \emph{et al.} \cite{pan} proposed PAN, which adopts a lightweight backbone and designs learnable post-processing. To complement the ability of the lightweight backbone to extract semantic features, they also proposed the feature pyramid enhancement module and feature fusion module. Liao \emph{et al.} \cite{db} introduced an adaptive threshold and added post-processing to the training, which improved detection performance with no extra computation. In addition, they utilized the Vatti clipping algorithm \cite{vatti1992generic} to simplify the expansion process further. Based on DBNet, ADNet \cite{ad} and RSMTD \cite{rmstd} adopted the prediction of the neural network to replace the computation of expansion distance, which improves performance but introduces an extra computation.  Although some methods achieve competitive performance and good speed, there is still room for improvement. Different from the above methods, the proposed similar mask maximally retains the geometric information of the text contour to improve the model performance. Moreover,  it is equipped with a super-simple post-processing which accelerated post-processing.

\subsection{Text Detection in Industrial and Traffic  Scene}
Industrial scene text detection is more challenging than natural scene, which enjoys low visual contrast and corroded surfaces. Guan \emph{et al.} \cite{rfn} collected an industrial text dataset and a synthetic industrial text dataset dubbed SynthMPSC. The former is the first industrial scene text detection dataset. In addition, to alleviate the inaccurate location problem, they proposed a refined feature-attentive network (RFN).

Different from other scenes, traffic scene text detection not only has a specific interference of motion blur but also has higher requirements for speed. Rong \emph{et al.} \cite{tgpd} established a challenging dataset for traffic scene detection, but training images are not available, and only 404 test images are available. To cope with it, Zhu \emph{et al.} \cite{zhu2017cascaded} collected a text-based traffic sign dataset. However,  this dataset is also not released. Moreover,  they adopted a fully convolutional network to predict candidate regions
of interest, followed by a lightweight network to detect text instances. Hou \emph{et al.} \cite{aam} proposed an attention anchor mechanism, which sets some predefined anchors and predicts weights for them to reduce the gap between ground truth and prediction. 
Based on DBNet,  Liang \emph{et al.} \cite{hfe} introduced a supervise for boundary and proposed a self-guided feature enhancement module to further improve detection performance. Although the above methods achieved competitive performance and detection speed, there is still significant room for improvement. To address these issues, we propose a new text representation method with a simple post-processing, which speeds up the inference process and achieves SOTA performance.

\section{Motion Blur Traffic Dataset}
\label{dataser}
Most public scene text datasets are mainly collected from natural scenes. Few traffic text datasets are established, such as Text-based Traffic Sign Dataset \cite{zhu2017cascaded}, Traffic Guide Panel Dataset (TGPT) \cite{tgpd}. However, the former is not publicly available, and the latter enjoys many labeled errors that only test sets are available. In this section,  we establish a motion blur traffic scene text (MBTST) dataset to promote the further development of traffic text detection.
\begin{table}[!t]
	\center
	
 {
		\caption{ Statistical comparison of the proposed dataset with other public datasets.  }
		\centering
		{ 
\begin{tabular}{ccccccc}
	
	\hline \multirow{2}{*}{ Dataset } & \multicolumn{2}{c}{ Image } & \multicolumn{2}{c}{ Label }    &\multirow{2}{*}{ Scene }  \\
	\cline{2-5}
	 & train & test   & word & line  \\
	 \hline
	 MPSC& 2555 & 639   & $\checkmark$ & -  &Industrial\\
	 MSRA-TD500 & 300 & 200   & - & $\checkmark$  &Natural\\
	ICDAR 2015 & 1000 & 500   & $\checkmark$ & - &Natural \\
	TGPD &- &404 & $\checkmark$ & - &Traffic\\
	ASAYAR-TXT &1000 &325 & $\checkmark$ & $\checkmark$ &Traffic\\
	  \hline
	MBTST(ours)& 1000 & 528   & $\checkmark$ & -  &Traffic\\
	\hline
	\label{datas}
\end{tabular}
}
}
\end{table}

\begin{figure}[t]
	\centering
	{
		\subfigure[Instance area distributions.]{
			
			\includegraphics[width=0.46\linewidth]{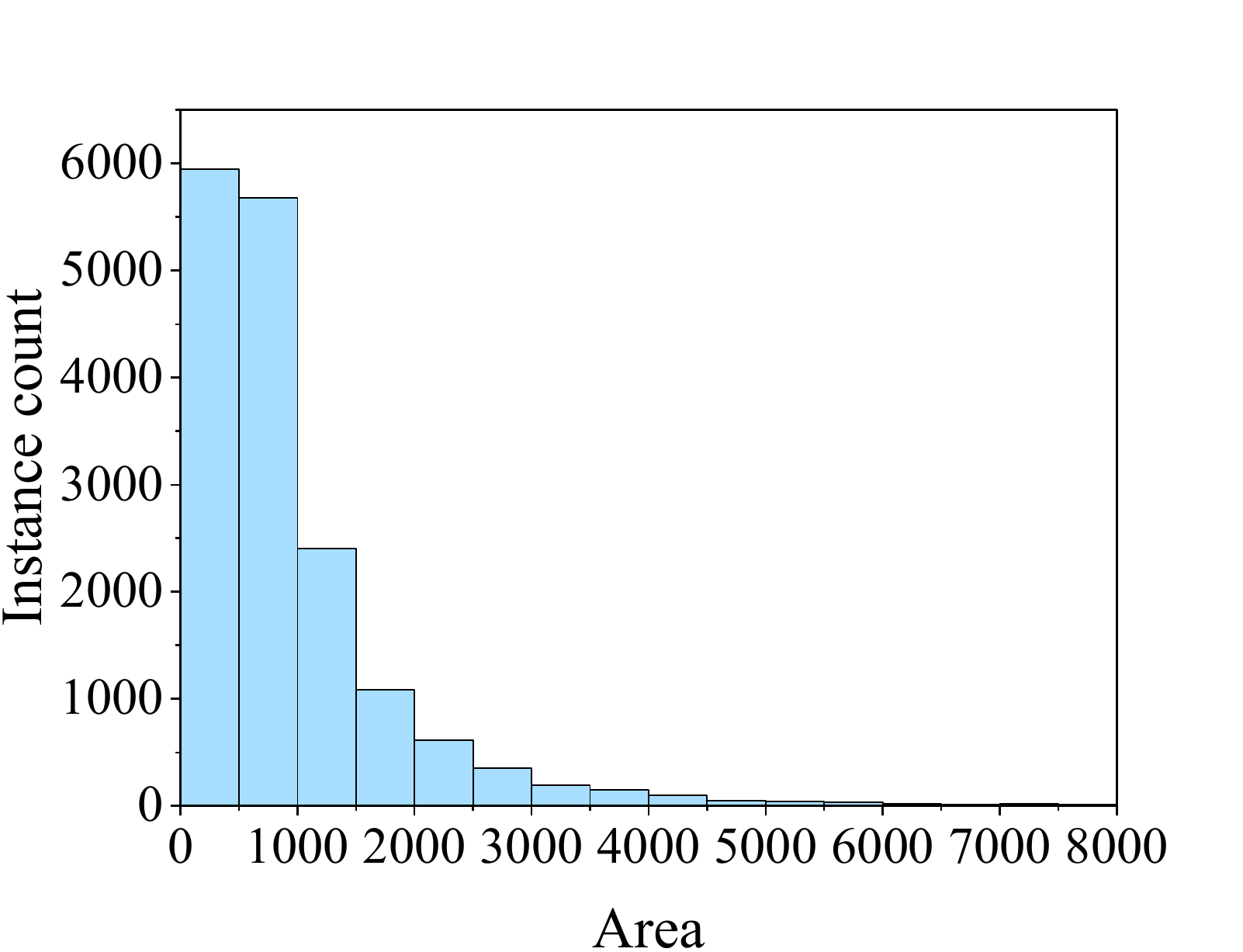}

		}\label{lim_a}
		\subfigure[Character  number distributions.]{

			\includegraphics[width=0.46\linewidth]{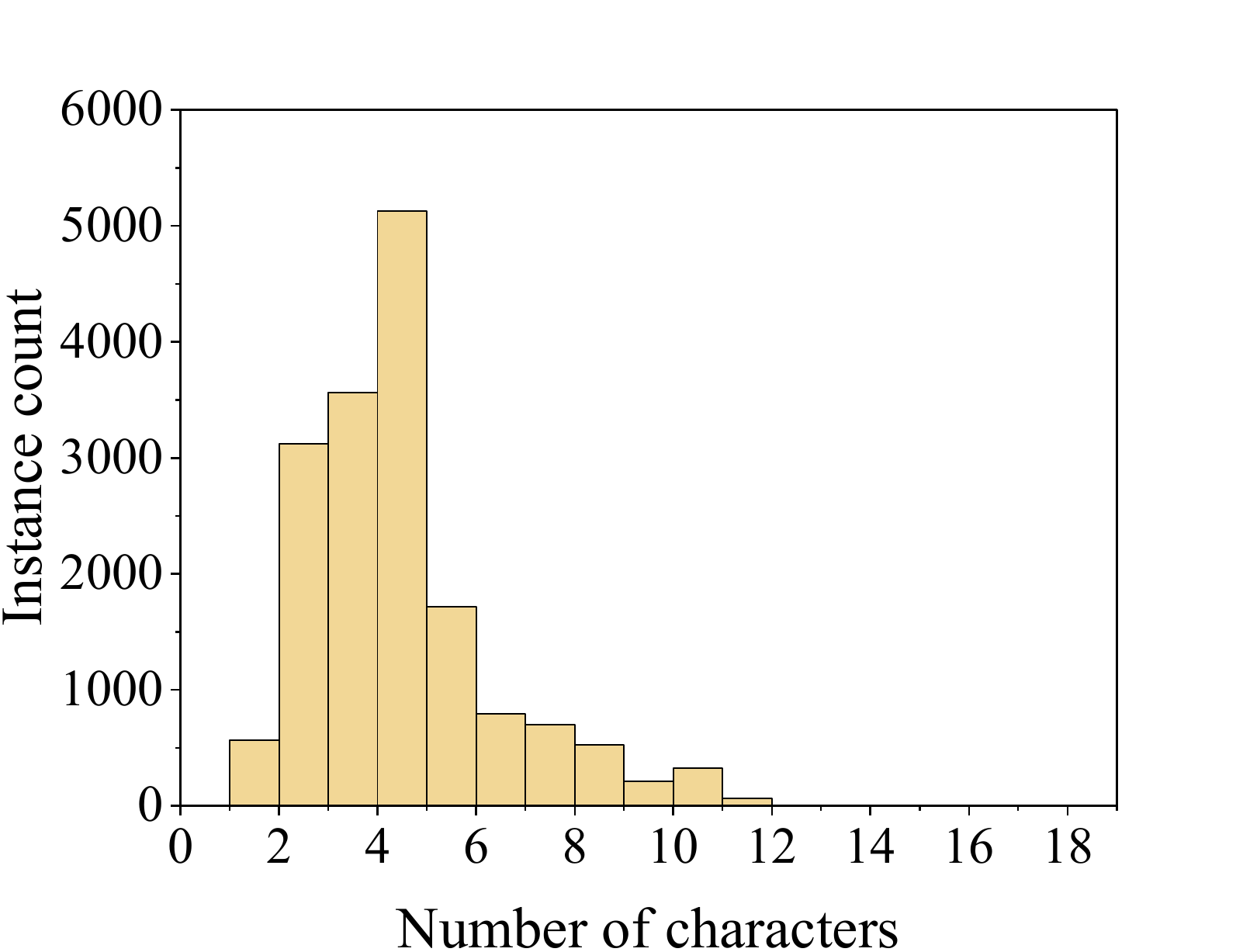}

		}\label{lim_b}}
	\centering

	\caption{The data distribution visualization of the proposed MBTST-1528 datasets. (a) Instance area distributions. (b) Character number distributions. }
	\label{tpa}
\end{figure}
\subsection{MBTST Dataset}
\subsubsection{Dataset Construction}
Data collection, cleaning, and labeling are performed to generate the proposed traffic scene text dataset. First, we collect traffic images from the web, and most of them include text. Then, we screen each image so that unqualified ones are removed. Next, follow the previous ICDAR2015 dataset \cite{karatzas2015icdar}, in which each instance in the images is labeled with four corner points. It is a word-level labeled dataset that is inspected in two rounds to reduce label errors.
\subsubsection{Dataset Analyze}
\begin{figure}[!t]
	\centering
	\includegraphics[width=1.0\linewidth]{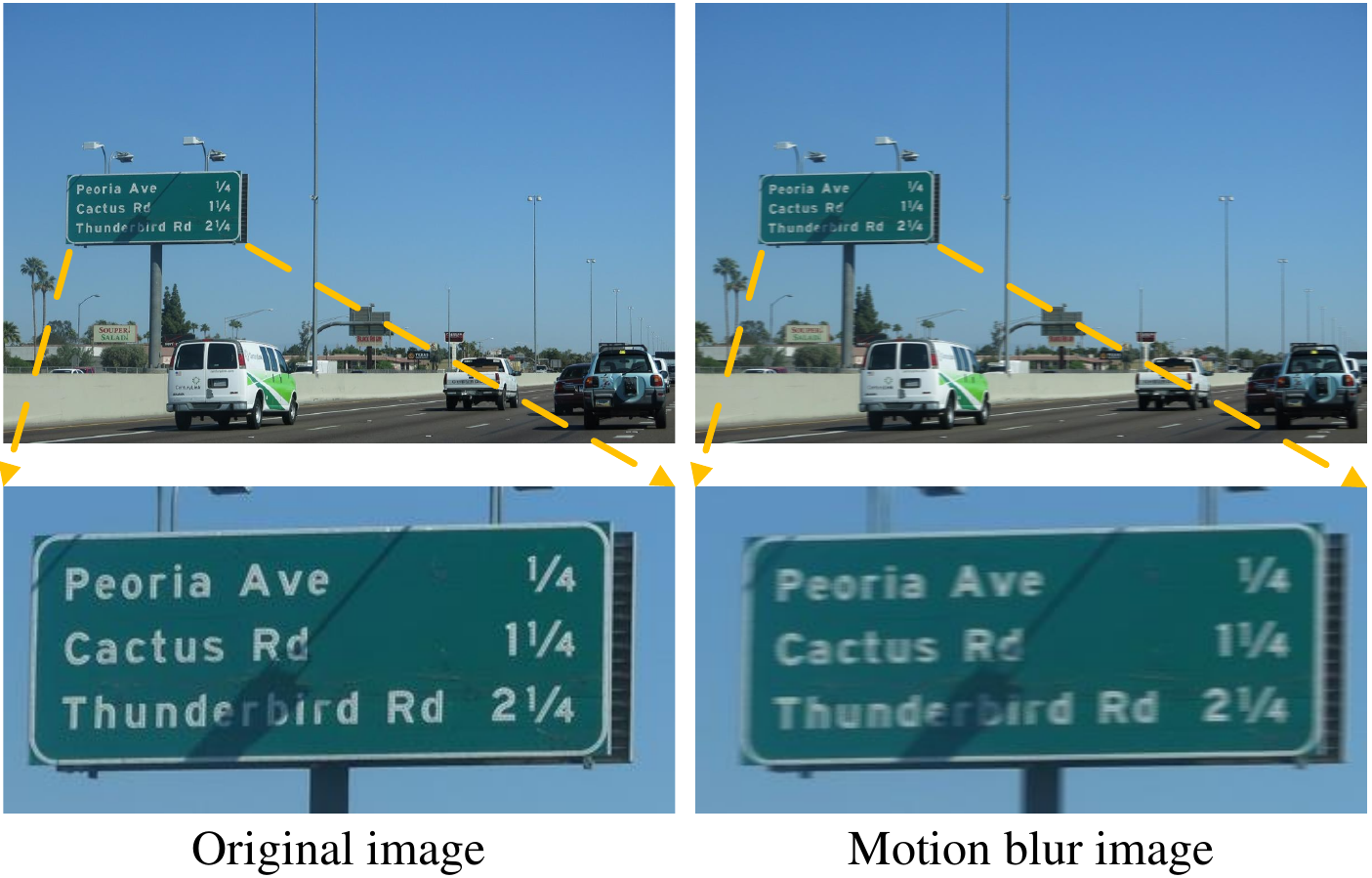}
	\caption{The visualization of original image and motion blur image.}
	\label{mbtsc_ab}
\end{figure}

The proposed dataset includes ground-truth boxes and text transcriptions.  In contrast to typical natural scene text, the majority of the text in traffic scenes deviates from regular words, often bearing specific semantics related to routes or toponymy. Illustrated in Fig \ref{tpa}, the area of the majority of instances is below 3000 pixels, equivalent to 0.3$\%$ of the total image pixels. Additionally, the majority of instances consist of 2 to 10 characters. In addition, each image introduces motion blur to simulate the real world, which is shown in Fig. \ref{mbtsc_ab}.
\subsection{Compared with other public datasets}
\label{data_compare}
Numerous widely used public benchmarks exist for text detection tasks, including ICDAR2015 \cite{karatzas2015icdar}, MSRA-TD500 \cite{yao2012detecting}, and MPSC \cite{}. The first two are derived from natural scenes, while the latter originates from industrial scenes. Regarding traffic scene datasets, the Text-based Traffic Sign Dataset \cite{zhu2017cascaded} and Traffic Guide Panel Dataset (TGPD) \cite{tgpd} have been proposed. Unfortunately, the former is not publicly accessible, and the latter suffers from numerous labeled errors, with only test sets available. In response, we collect and establish the MBTST dataset for traffic scene text detection. In addition, there are also some less-used datasets. Additionally, there are less commonly used datasets. The corresponding details are presented in Table \ref{datas}.

\begin{figure}[!t]
	\centering
	\includegraphics[width=1.0\linewidth]{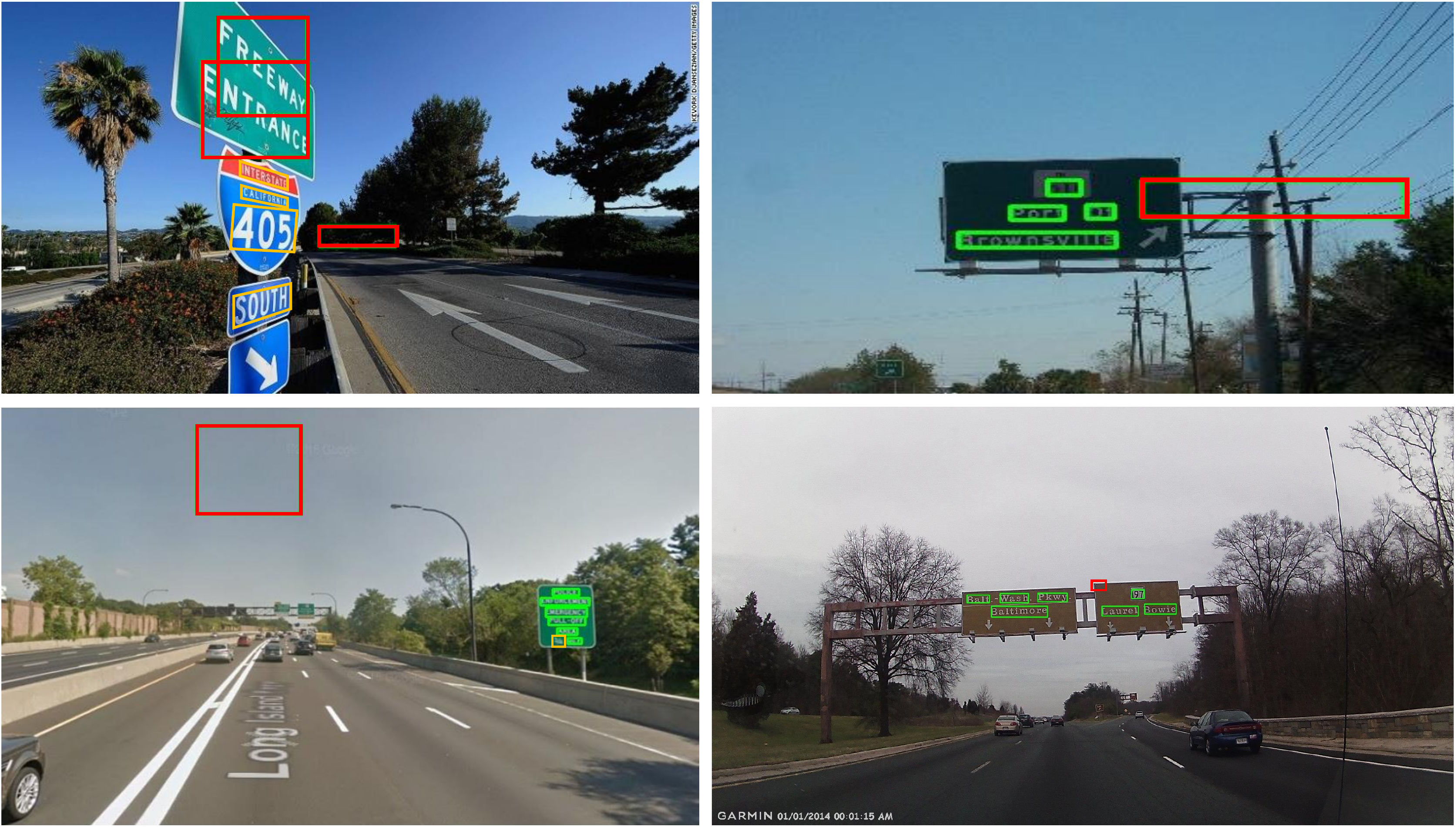}
	\caption{The visualization of some error annotation on the TGPD dataset. The error and missing annotations are labeled in red and orange, respectively. }
	\label{tgpt}
\end{figure}
\subsubsection{MPSC} It is an industrial text dataset that presents additional challenges compared to natural scenes, including uneven light, low contrast, and unsmooth surfaces. It comprises 2,555 training images and 639 testing images.

\subsubsection{ICDAR2015}
It includes numerous images from supermarkets with low resolution. It comprises 1,000 training images and 500 testing images, and each instance is labeled with four corner points.
\subsubsection{MSRA-TD500}
A bilingual text detection dataset featuring English and Chinese texts. Each instance is labeled at the line level. With only 300 training images, previous methods often incorporate HUST-TR400 \cite{yao2014unified} for training. It only has 200 testing images.
\subsubsection{TGPD}
It is a dataset for traffic guide panel recognition featuring thousands of traffic texts. Unfortunately, only the 404 test images are available, and the corresponding annotations contain numerous errors. It is shown in Fig. \ref{tgpt}.
\subsubsection{ASAYAR-TXT } ASAYAR \cite{asayar} consists of three sub-datasets. The ASAYAR-TXT mainly contains texts collected from Moroccan highways. Each instance is labeled at the word level and line level. It contains 1,500 images, 1,100 for training and 275 for testing.
\subsubsection{CRPD \textnormal{\cite{crpd}}} It is a large public Chinese License Plate dataset, which is mainly used to detect and recognize license plates. It is mainly collected from different provinces of China in different weathers. It also includes tree sub-datasets, which are divided according to the number of objects. The largest sub-dataset is the CRPD-single, which has 20k images for training, 5k images for validation, and 1k images for testing. The another sub-dataset CRPD-multi includes 1,000 training images, 250 validation images, and 300 testing images. 
\subsubsection{MBTST-1528}
It is a traffic text dataset collected from web by ours. It includes 1,000 images for training and 528 for testing, many collected from the highway. The motion blur is given full consideration in this dataset, which is inspected two times to reduce the labeled errors. The original image and motion blur image are visualized in Fig. \ref{mbtsc_ab}. 
\section{Method}
\label{method}
In this section, we present an effective and efficient multi-scene text detector. We start by providing a detailed illustration of the similar mask and feature correction module. Subsequently, we outline the whole pipeline of the proposed method. Finally, we introduce the employed loss function.

\begin{figure}[!t]
	\centering
	\includegraphics[width=1.0\linewidth]{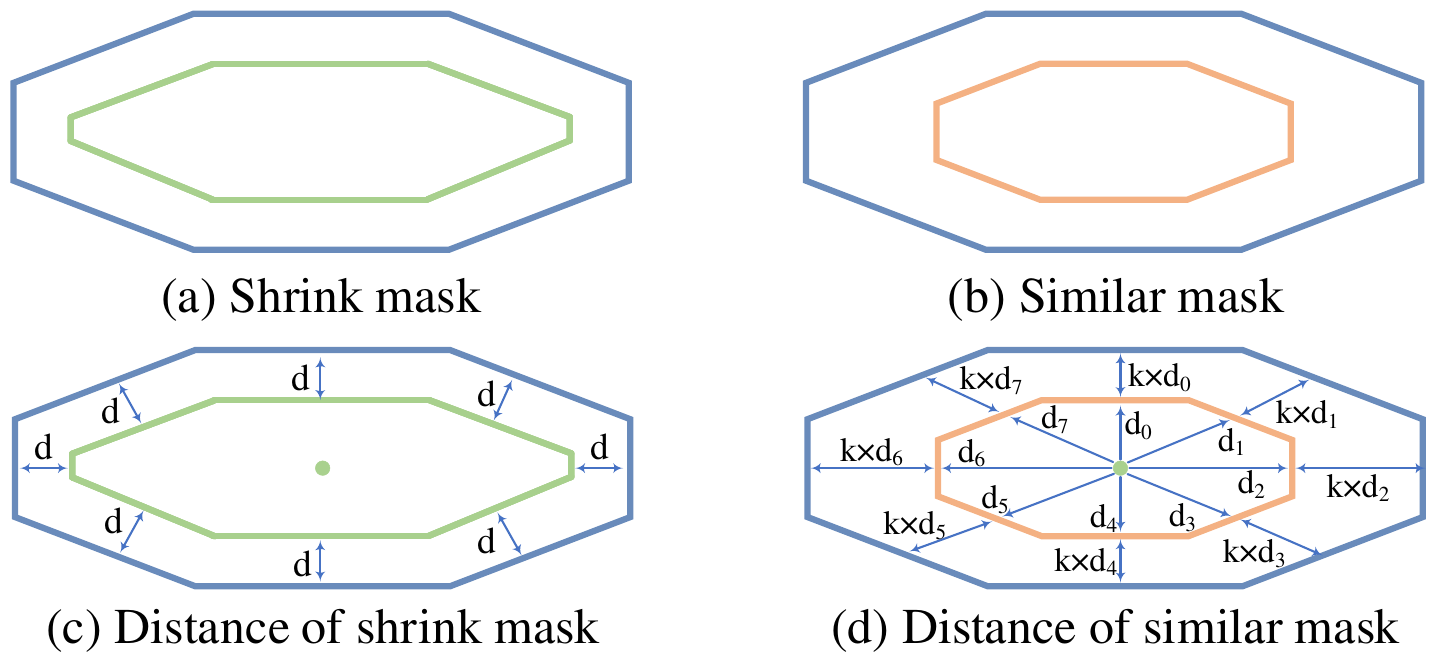}
	\caption{The label generation visualization of shrink mask and similar mask. The blue, green, and orange polygons represent text contour, shrink mask, and similar mask, respectively. }
	\label{vis_shrink}
\end{figure}
\begin{figure}[!t]
	\centering
	\includegraphics[width=1.0\linewidth]{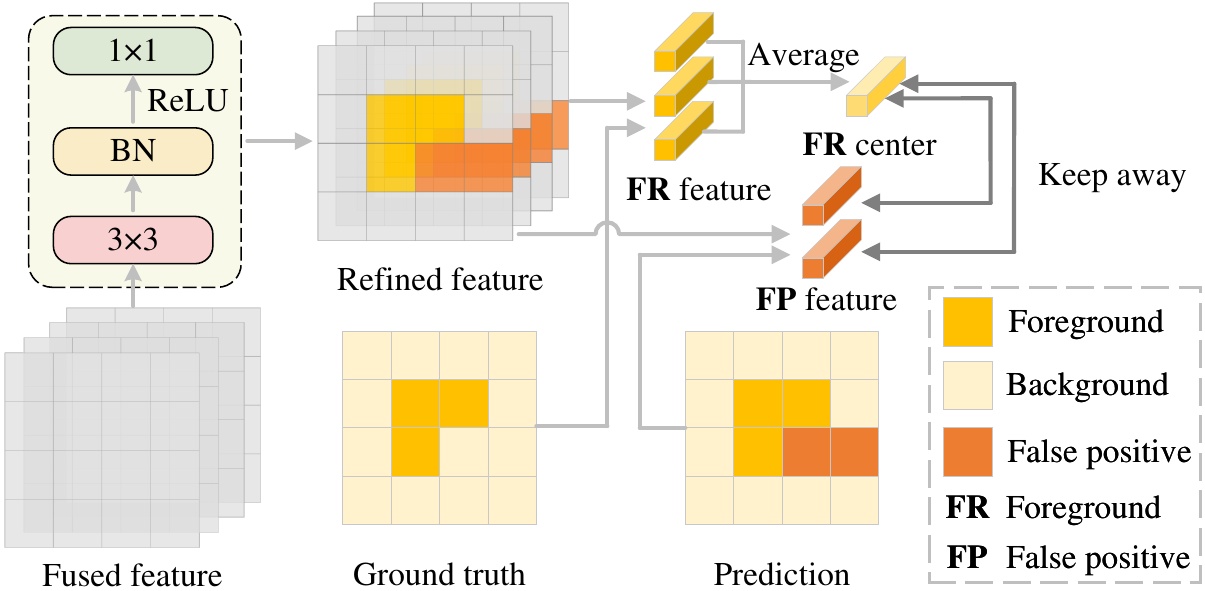}
	\caption{The visualization of feature correction module.}
	\label{fcm}
\end{figure}
\subsection{Similar Mask}
The majority of segmentation-based methods employ a shrink mask to reconstruct text instances, achieving SOTA performance with rapid detection speed. The shrink mask is calculated by shrinking the instance contour with a specific distance $d$, which can be computed as follows:
\begin{equation}
	d = \frac{S\times(1-\gamma^2)}{L},
\end{equation}
where $S$ and $L$ represent the area and perimeter of the instance, respectively. $\gamma$ represents the shrink coefficient, signifying the degree of contraction. The shrink mask and corresponding text contour are visualized in Fig. \ref{vis_shrink} (a) and Fig. \ref{vis_shrink} (c).
During the testing stage, the predicted shrink mask is initially binarized. Subsequently, contour extraction is executed. As a result, the contour of the shrink mask is expanded by a distance $d_e$, calculated as follows:  
\begin{equation}
	d_{e} = \frac{S^{\prime}\times \beta}{L^{\prime}},
\end{equation}
where $S^{\prime}$ and $L^{\prime}$ represent the area and perimeter of the shrink mask. $\beta$ is the extended coefficient. 

\begin{figure*}[!t]
	\centering
	\includegraphics[width=1.0\linewidth]{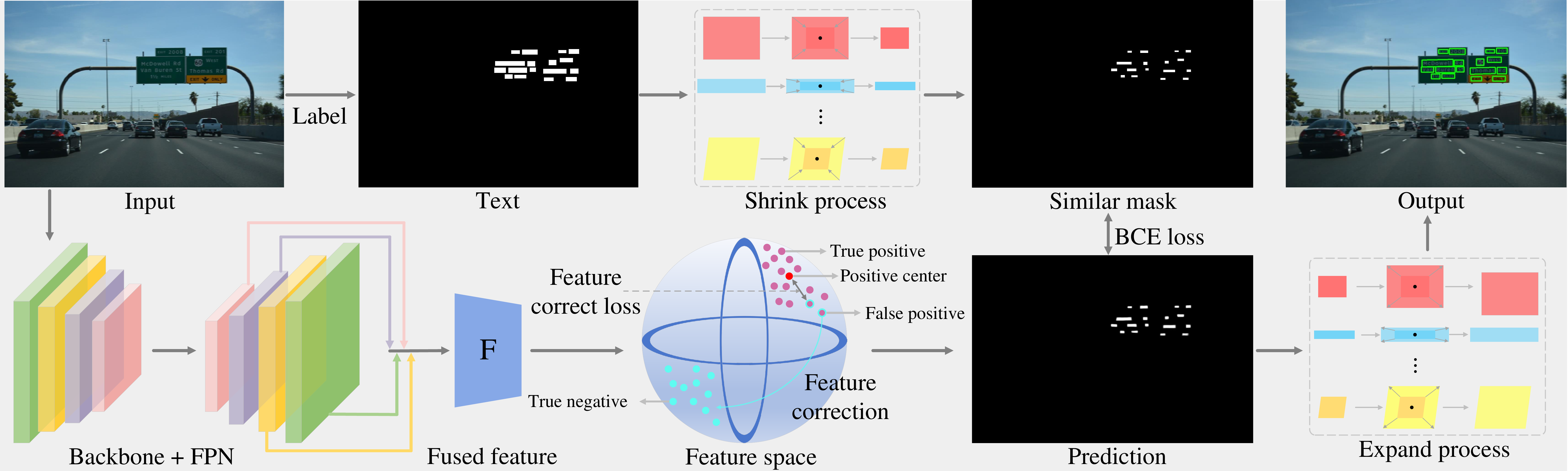}
	\caption{The whole pipeline of the proposed SMNet, which includes the label generation procedure, backbone, FPN, feature correction module, and a simple post-processing. }
	\label{overview}
\end{figure*}

The Similar Mask exhibits two significant limitations: 1) Due to both area and perimeter being global features, it overlooks local features and loses some important geometry information. 2) Its post-processing involves complex operations, including area and perimeter calculations and shrink contour expansion, significantly impacting detection speed. For instance, in DBNet \cite{db}, post-processing consumes approximately 30$\%$ of the test time.

Diverging from the shrink mask, we introduce a novel text representation method called Similar Mask, aiming for both efficiency and effectiveness. Specifically, it endeavors to preserve the features of the text contour to the maximum extent. Moreover, it incorporates an ultra-simplified post-processing step, significantly enhancing detection speed.
Considering a text contour, which can be denoted by a series of points ${P_1, P_2, ..., P_{n}}$, the text center of the instance is calculated as:
\begin{equation}
	t_{c} = \frac{\sum_{i=1}^n P_{i}}{n}. \quad i=1,2, ..., n
\end{equation}
The above points are expressed in the form of horizontal and vertical coordinates.
The similar mask is also consisted of a series of points ${P_1^s, P_2^s,  ...,  P_{n}^s}$, which are generated by:
\begin{equation}
	d_{i} = P_i - t_c, \quad i=1,2, ..., n
\end{equation}
\begin{equation}
	P_{i}^s = t_c + d_i \times \delta, \quad i=1, 2, ..., n
\end{equation}
where $\delta$ represent the shrink parameter. The similar mask and corresponding text contour are shown in Fig. \ref{vis_shrink} (b) and Fig. \ref{vis_shrink} (d). In order to provide a intuitive representation, we substitute "$\delta$" with "$k$" in the figure, where the relationship between  "$\delta$" with "$k$" is as follows:
\begin{equation}
	k = \frac{ 1- \delta}{ \delta}.
\end{equation}
In the inference phase, the similar mask is binarized and extracted contour first to get a series of points ${\hat{P}_1^{s}, \hat{P}_2^{s}, ...,  \hat{P}_{n}^{s}}$. The center point $\hat{t}_c$ of the similar mask is calculated as:
\begin{equation}
	\hat{t}_c = \frac{\sum_{i=1}^n \hat{P}_{i}^{s}}{n}.
\end{equation}
The text contour points $\hat{P}$ are computed as:
\begin{equation}
	\hat{d}_{i} = \hat{P}_{i}^s - \hat{t}_c, \quad i=1,2, ..., n
\end{equation}

\begin{equation}
	\hat{P}_{i} = \hat{t}_c + \hat{d}_{i} \times \frac{1}{\delta }. \quad i=1, 2, ..., n
\end{equation}
The text contour points $\hat{P}_{1}, \hat{P}_{2}, ..., \hat{P}_{n}$ are connected to form the text contour, which is the final detection result.

\subsection{Feature Correction Module} Previous segmentation-based methods only optimize the module by judging whether the pixel is text. Directly predicting this target is challenging, as effectively transferring knowledge between input and output is difficult. Relying solely on output to guide the model is insufficient. Hence, we introduce the feature correction procedure to guide the module at the feature level. It assists the model in recognizing text features and distinguishing between foreground and background. Specifically, we use a Similar Mask to reconstruct the text contour, an artificially defined geometric concept with incomplete semantic features that particularly require this knowledge's assistance. The corresponding details are illustrated in Fig. \ref{fcm}. Initially, the fused feature undergoes smoothing to generate the refined feature, described as follows:
\begin{equation}
  	\textrm{H}= (\textrm{Conv}_{3\times3}\_ \textrm{BN} \_ \textrm{ReLU}(\textrm{F})),
\end{equation}
\begin{equation}
\textrm{F}_r= (\textrm{Conv}_{1\times1}(\textrm{H})),
\end{equation}
where $\rm F$, $\rm H$, and $\textrm{F}_r$ represent the fused feature, hidden feature, and refined feature, respectively. The ground truth and prediction are down-sample through max-pooling to keep the size the same with the refined feature map. According to the ground truth, the position of the foreground pixel is obtained. The corresponding foreground feature is averaged to compute the foreground feature center. It can be described as follows: 
\begin{equation}
	\rm	\textrm{G}^{'}  =  \textrm{Maxpool}(\textrm{G}),
\end{equation}
\begin{equation}
	{\textrm{F}_g(j)} =
	\begin{cases}
		\textrm{F}_r(j),  &{\text{if}}  \quad \textrm{G}^{'}(j) = 1,\\
		{0,}  &{\text{otherwise.}}
	\end{cases}
\end{equation}
\begin{equation}
		\textrm{F}_g^{'} = \frac{\sum_{i=1}^N \textrm{F}_g(j) }{N} , {\textrm{F}_g(j)} \neq 0,
\end{equation}
where $\rm G$ and $\rm G^{'}$ represent the ground truth and ground truth after downsampling.
$\textrm{F}_g$ is the filterd feature and $\textrm{F}_g^{'}$ is the corresponding groud truth feature center, where $N$ is the pixel number of the positive ground truth. Samely, the false positive feature is obtained as:

\begin{equation}
	\textrm{P}^{'}  =  \textrm{Maxpool}(\textrm{P}),
\end{equation}
\begin{equation}
	{\textrm{F}_{fp}(j)} =
	\begin{cases}
		\textrm{F}_r(j),  &{\text{if}}  \quad \textrm{G}^{'}(j) = 0 ~ \textrm{and} ~ \textrm{P}^{'}(j) > \theta,\\
		{0,}  &{\text{otherwise.}}
	\end{cases}
\end{equation}
where $\textrm P$ and $\textrm P^{'}$ represent the prediction and prediction after downsampling. $\textrm{F}_{fp} \neq 0$ is the set of false positive feature.
The traget of feature correct module is keep away  $	 {\textrm{F}_{fp}(j)} \neq 0$ with $\textrm{F}_g^{'}$ for any pixel $j$.

\subsection{Overall Pipeline}
In this paper, a real-time multi-scene text detector is proposed, which speeds up the post-processing and improves the detection accuracy significantly. The overall structure is shown in Fig. \ref{overview}, which includes the process of similar mask generation, backbone, feature pyramid network (FPN) \cite{fpn}, similar mask prediction header, and feature correct module. First, we used the given text annotation to generate 
a similar mask label.   ResNet \cite{resnet} with deformable convolution \cite{deformable2} is adopted as the backbone. It is utilized to extract different scale feature maps, whose sizes are $\frac{1}{4}$, $\frac{1}{8}$, $\frac{1}{16}$, and $\frac{1}{32}$ of the input images. Then, the FPN is used to fuse these feature maps. The similar mask prediction layer is the segmentation head that is used to predict the similar mask. It consists of one convolution layer and two transposed convolution layers, which upsample the feature map from $\frac{H}{4} \times  \frac{W}{4} $ enlarge to $ H \times W$,  where $W$ and $H$ are the width and height of the input image. The feature correction module is a training-only operation, which can be removed during the test stage. In the end, simplified post-precessing is used to reconstruct the text instance contour to obtain final detection results.
\begin{table*}[!t]
	\center
	
{
		\caption{  Ablation study on the effect of shrink mask, similar mask, and feature correction module on detection performance on the  MSRA-TD500 and MBTST-1528.  The ``Shrink'' and ``Similar'' represent shrink mask and similar mask, respectively}
		\centering
		{ 
			\begin{tabular}{ccc|cccc|cccc}
				
				\hline
				\multirow{2}{*}{Shrink}  &\multirow{2}{*}{Similar} 
				&\multirow{2}{*}{FCM} 
				& \multicolumn{4}{c}{MSRA-TD500}	        &\multicolumn{4}{c}{MBTST-1528}  	  \\    
				\cline{4-11} 
				&		& &Precision &Recall &F-measure &FPS &Precision &Recall &F-measure &FPS \\ \hline

				\checkmark &$\times$ &$\times$  &86.4 &78.5 &82.3  &54.2 &94.1 &83.8 &88.7&59.0 \\ 
				$\times$ &	\checkmark  &$\times$  &86.6 &78.9 &82.6&61.5  &94.4 &84.2 &89.0 &69.1  \\
				\checkmark &$\times$  &	\checkmark   &88.8 &79.4 &83.8 &53.1   &95.1 &84.5 &89.5 &59.3  \\ 
				$\times$ &	\checkmark   &	\checkmark  &88.9 &81.3 &84.9  &62.0 &96.7 &84.2 &90.0 &69.8  \\
				\hline
			\end{tabular}
			
		}
		\label{tab_abs1}	
		
	}
\end{table*}

\subsection{Optimization}
In this paper, three losses are utilized to optimize the proposed method. To be specific, they are similar mask prediction loss $\mathcal{L}_{s}$,  and feature correction loss $\mathcal{L}_{fc}$.
The binary cross-entropy (BCE) loss is adopted for $\mathcal{L}_{s}$. To alleviate the unbalance of the positive and negative samples, hard negative mining is applied in the BCE loss, which can be described as follows:

\begin{equation}
		\mathcal{L}_{s}= \sum\limits_{i\in S_{l}}-y_i \times log(\hat{y}_i)-(1-y_i) \times log(1-\hat{y}_i),
\end{equation}
where $y_i$ and $\hat{y}_i$ represent the ground truth and prediction of the shrink mask. $S_l$ is the selected sample set that the ratio of positives and negatives is 1:3. 

For the feature correction loss, cosine similarity is used, which can be computed as follows:
\begin{equation}
		Cos(\textrm{F}_{fp}(j) ,\textrm{F}_g^{'}) =  \frac{ \textrm{F}_{fp}(j) \times  \textrm{F}_g^{'}}{|\textrm{F}_{fp}(j)| \times| \textrm{F}_g^{'}|} , 
\end{equation}
\begin{equation}
		\mathcal{L}_{fc} = \frac{1}{M} \times \sum_{j=1}^M \frac{1}{1 + e^{-\sigma \times Cos(\textrm{F}_{fp}(j) ,\textrm{F}_g^{'})}} , 
\end{equation}
where $F_{fp}$ and $F_g^{'}$ represent the false positive feature and ground truth feature center. 

The overall loss function during the training can be represented as:
\begin{equation}
	\mathcal{L}= \lambda_{1}\mathcal{L}_{s}+\lambda_{2}\mathcal{L}_{fc},
\end{equation}
where $\lambda_{1}$ and $\lambda_{2}$ are the weight of losses to balance the multi-task of the module. 

\section{Experiments}
\label{experiments}
In this section, we first introduce the dataset used and the evaluation metrics adopted. Then, the corresponding implementation details are described. Subsequently, we perform ablation experiments on two datasets to verify the effectiveness of the proposed method. Furthermore, the detection performance on multiple datasets proves the superiority of the proposed SM-Net. In addition, the detection results on multiple scenes are visualized.

\subsection{Datasets and Evaluation Metrics}
Most of the used datasets are introduced in the section \ref{data_compare}. SynthText \cite{gupta2016synthetic} is a synthetic dataset that consists of 800k images. It includes various texts and different scenes, which is used to pre-train to improve the robustness of the module.

To ensure a fair comparison, we adopted the precision (P), recall (R), F-measure(F), and frames per second (FPS) as the evaluation metrics follow the previous works. The F-measure is computed by the precision and recall, which is used to represent the detection performance. FPS is used to represent the detection efficiency. It can be formulated as follows:
\begin{equation}
	P = \frac{TP}{TP+FP}\times100\%,
\end{equation}
\begin{equation}
R = \frac{TP}{TP+FN}\times100\%,
\end{equation}
\begin{equation}
	F = \frac{2\times P \times R}{P +R }\times100\%,
\end{equation}
where $TP$, $FP$, and $FN$ represent the true positive, false positive, and false negative detection. 

\subsection{Implementation Details}
\begin{table*}[!t]
	\center
	
	{
		\caption{  Ablation study on the effect of shrink and extend coffecient on detection performance on the ICDAR2015 and MSRA-TD500.  }
		\centering
		{ 
			\begin{tabular}{c|cc|ccc|ccc}
				
				\hline
				&\multirow{2}{*}{Shrink}  &\multirow{2}{*}{Extend} 
				& \multicolumn{3}{c}{MSRA-TD500}	        &\multicolumn{3}{c}{ICDAR2015}  	  \\    
				\cline{4-9} 
				&	&	 &Precision &Recall &F-measure  &Precision &Recall &F-measure  \\ \hline 
				
				\multirow{6}{*}{SM-Net with}	
				&0.4 &2.5   &87.2 &78.5 &82.6   &89.2 &78.4 &84.0 \\ 
				&0.5 &2    &87.5 &80.8 &84.0  &88.3 &79.7 &83.8   \\
				&0.55 &1.818   &87.8 &80.6 &84.1    &87.5 &80.8 &84.0  \\ 
				&0.6 &1.667   &88.9 &81.3 &84.9  &88.0 &79.4 &83.5   \\
				&0.65 &1.538   &87.5 &79.0 &83.0   &87.2 &79.5 &83.2   \\ 
				&0.7 &1.429   &87.3 &80.2 &83.6  &86.7 &78.4 &82.4  \\
				\hline
			\end{tabular}
			
		}
		\label{tab_abs2}	
		
	}
\end{table*}

\begin{table}[t]
	\center

	{
		\caption{ Comparison with existing state-of-the-art (SOTA) approaches on the MSRA-TD500 dataset. These methods are divided into real-time and non-real-time methods according to the detection speed. ``\textcolor{blue}{\textbf{Blue}}'' and ``\textcolor{red}{\textbf{Red}}'' represent the  best performance of real-time and no-real-time methods, respectively. }
		\setlength{\tabcolsep}{6pt}
		\begin{tabular}{c|c|cccc}

			\hline Methods &Venue  & P & R & F & FPS \\
			\hline 
			Yao \emph{et al.}\cite{yao2012detecting} & CVPR'12 &63 &63 &60 &0.14 \\
			TexStar\cite{texstar} & TPAMI'14 &71 &61 &66 &1.25 \\ \hline
			PAN\cite{pan}  &ICCV'19 &84.4 &83.8 &84.1 &30.2 \\ 
			DBNet\cite{db}  &AAAI'20 &{90.4} &{76.3} &{82.8} &62 \\
			CTNet\cite{ct}  &NeurIPS'21 &90.0 &82.5  &86.1 &34.8\\ 
			PAN++\cite{pan++} &TPAMI'22  &85.3 &84.0 &84.7 &32.5 \\ 
			CMNet\cite{cm} &TIP'22 &89.9 &80.6 &85.0 &41.7 \\
			ZTD \cite{zoom} &TNNLS'23  &91.6 &82.4 &86.8 &59.2\\
			DBNet++\cite{db++}  &TPAMI'23 &87.9 &82.5 &85.1 &55 \\	
			RSMTD \cite{rmstd} &TMM'23  &89.8 &83.1 &86.3 &62.5\\
			SMNet-Res18 &Ours   &{89.9} &{85.6} &\textcolor{blue}{\textbf{87.7}} &62.8 \\ \hline

			CRAFT\cite{craft} &CVPR'19  &88.2 &78.2 &82.9 &8.6  \\  
			DBNet\cite{db} &AAAI'20 &{91.5} &{79.2} &{84.9} &{32}\\ 
			
			LEMNet\cite{lem} &TMM'22  &85.6 &84.8 &85.2 &-\\
			ADNet\cite{ad} &TMM'22  &92.0 &83.2 &87.4 &-\\
			EMA\cite{ema} &TIP'22 &88.7 &81.1 &84.7 &24.2 \\
			DBNet++\cite{db++} &TPAMI'23 &{91.5} &{83.3} &{87.2} &{29} \\ 
			RP-Text \cite{rp} &TMM'23 &88.4 &84.6 &86.5 &-\\		
			HFENet \cite{hfe} &TITS'23 &98.8 &84.0 &88.2 &21.4\\ 
			MorphText\cite{morph} &TMM'23	 &90.7 &83.5 &87.0 &-\\
			VRRCA\cite{vrrca} &TMM'23 &90.5 &83.8 &87.0 &- \\
			FS \cite{fs} &TIP'23 &89.3 &81.6 &85.3 &25.4 \\
			LeafText\cite{leaftext} &TMM'23  &92.1 &83.8 &87.8 &-\\

			RFN\cite{rfn} &TCSVT'22 &88.4 &87.8 &88.1 &- \\

			LOAD\cite{load} &TCSVT'23 &88.8 &86.1 &87.4 &-\\
			SMNet-Res50 &Ours  &{91.0} &{86.8} &\textcolor{red}{\textbf{88.8}} &23.1\\

			\hline
		\end{tabular}
		\label{td500}
	}
	
\end{table}

The backbone is ResNet with deform convolution, which is pre-trained on ImageNet. Then, we adopt a feature pyramid network (FPN) to fuse different scale features. The whole network is pre-trained on the SynthText dataset. The proposed SM-Net is trained in 1200 epochs with an initial learning rate of 0.007. The learning rate is adjusted by SGD and the ``poly'' strategy. During the training stage, all the images are resized to 640$\times$640. In addition, random cropping, flipping, and rotation are utilized for data augmentation. The loss coefficients $\lambda_1$ and $\lambda_2$ are set to 6 and 0.02, respectively. All the detection results are tested on a GTX 1080Ti with a single i7-6800K.

\subsection{Ablation Study}
A series of ablation studies are conducted on the MSRA-TD500 and MBTST-1528. Note that no additional datasets are used to pre-train for a fair comparison. Compared with the baseline, the proposed SM-Net improves the detection performance and efficiency.

\subsubsection{Influence of the similar mask} Most shrink-based methods usually use shrink masks to represent text instances. We proposed an effective and efficient representation method to promote further text detection research, which enjoys simplified post-processing. As shown in the table \ref{tab_abs1}, the proposed similar mask brings 0.3$\%$ and 0.3$\%$ gains in F-measure on the MSRA-TD500 and MBTST-1528 datasets. When equipped with the FCM, the improvements become 1.1$\%$ and 0.5$\%$, respectively. Although the detection improvement is limited, the inference speed is 1.2 times faster than the baseline. Compared to the shrink mask, the scheme used the similar mask saves 50$\%$ of post-processing time, which proves the superiority of the proposed similar mask.
\begin{figure}[!t]
	\centering
	\includegraphics[width=0.95\linewidth]{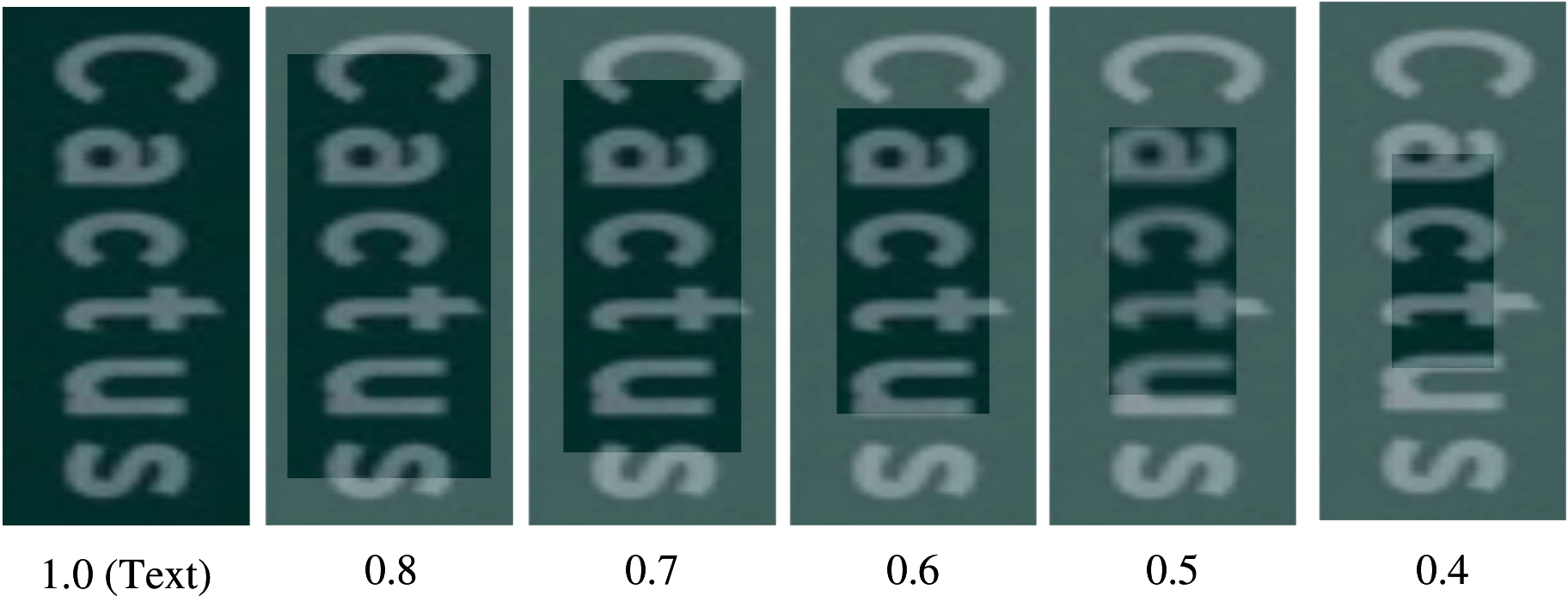}
	\caption{The visualization of the similar mask under different shrink cofficient.}
	\label{ratio}
\end{figure}

\begin{table}[t]
	\center	{	
		\caption{ Comparison with traditional methods on the MSRA-TD500 dataset.  }

		\begin{tabular}{c|cccc}

			\hline Methods   & Precision & Recall & F-measure & FPS \\
			\hline 
			Yao \emph{et al.}\cite{yao2012detecting}  &63 &63 &60 &0.14 \\
			Yin \emph{et al.}\cite{texstar}  &71 &61 &66 &1.25 \\
			Kang \emph{et al.}\cite{Kang_2014_CVPR}  &71 &62 &66 &- \\
			Yao \emph{et al.}\cite{yao2014unified} &64 &62 &61 &0.26 \\
			Yin \emph{et al.}\cite{momv}  &81 &63 &71 &0.71 \\
			 \hline
			SMNet    &{89.9} &{85.6} &87.7 &62.8 \\ 
			
			\hline
		\end{tabular}
		\label{td500_trad}
	
}
\end{table}

\subsubsection{Effectiveness of the feature correction module} Previous methods usually adjust the model according to the prediction. To pursue a superior model further, the feature correction module (FCM) is proposed to refine the module at the feature level. As shown in Table \ref{tab_abs1}, when using the shrink mask to represent the text instance, FCM brings  1.5$\%$ and 0.8$\%$ improvement in F-measure on MSRA-TD500 and MBTST-1528, respectively. For the proposed similar mask, the proposed archives 1.6$\%$ and 1.6$\%$ gains on MSRA-TD500 and MBTST-1528. Furthermore, the FCM is a training-only operation, which can be removed during the test stage and introduces no extra computation. In addition, FCM is a plug-and-play module that can insert existing modules.
\subsubsection{Influence of the shrink coffecient $\delta$}
We conduct experiments on MSRA-TD500 and ICDAR2015 to analyze the performance under different shrink coefficients. The similar mask under different shrink coefficients is visualized in Fig. \ref{ratio}. As shown in Table \ref{tab_abs2}, SMNet achieves the best performance at 88.9$\%$, 81.3$\%$, and 84.9$\%$ in terms of precision, recall, and F-measure when the $\delta$ is set to 0.6 for the MSRA-TD500. For the ICDAR2015 dataset, the best scheme is set $\delta$ to 0.5 or 0.4, which achieves F-measure in 84.0 $\%$. Note that, with the continued increase of $\delta$, the performance begins to decrease. These experiments show the different appropriate settings for line-level and word-level datasets and provide consultations for other datasets.

\begin{figure}[!t]
	\centering
	\includegraphics[width=0.95\linewidth]{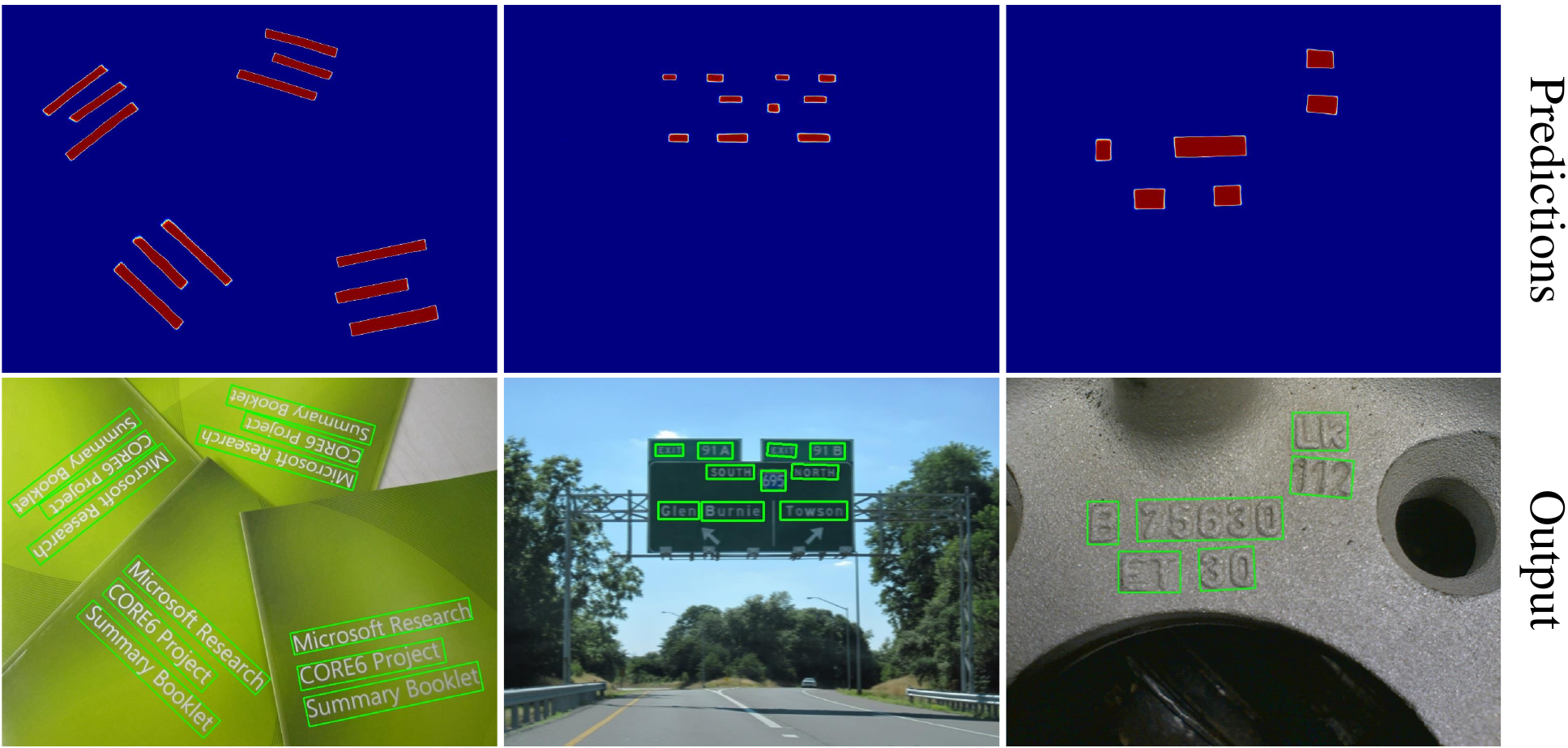}
	\caption{The visualization of the prediction and corresponding outputs.}
	\label{sm}
\end{figure}
\subsection{Comparision with existing SOTA methods}
\begin{figure*}[!t]
	\centering
	\includegraphics[width=1.0\linewidth]{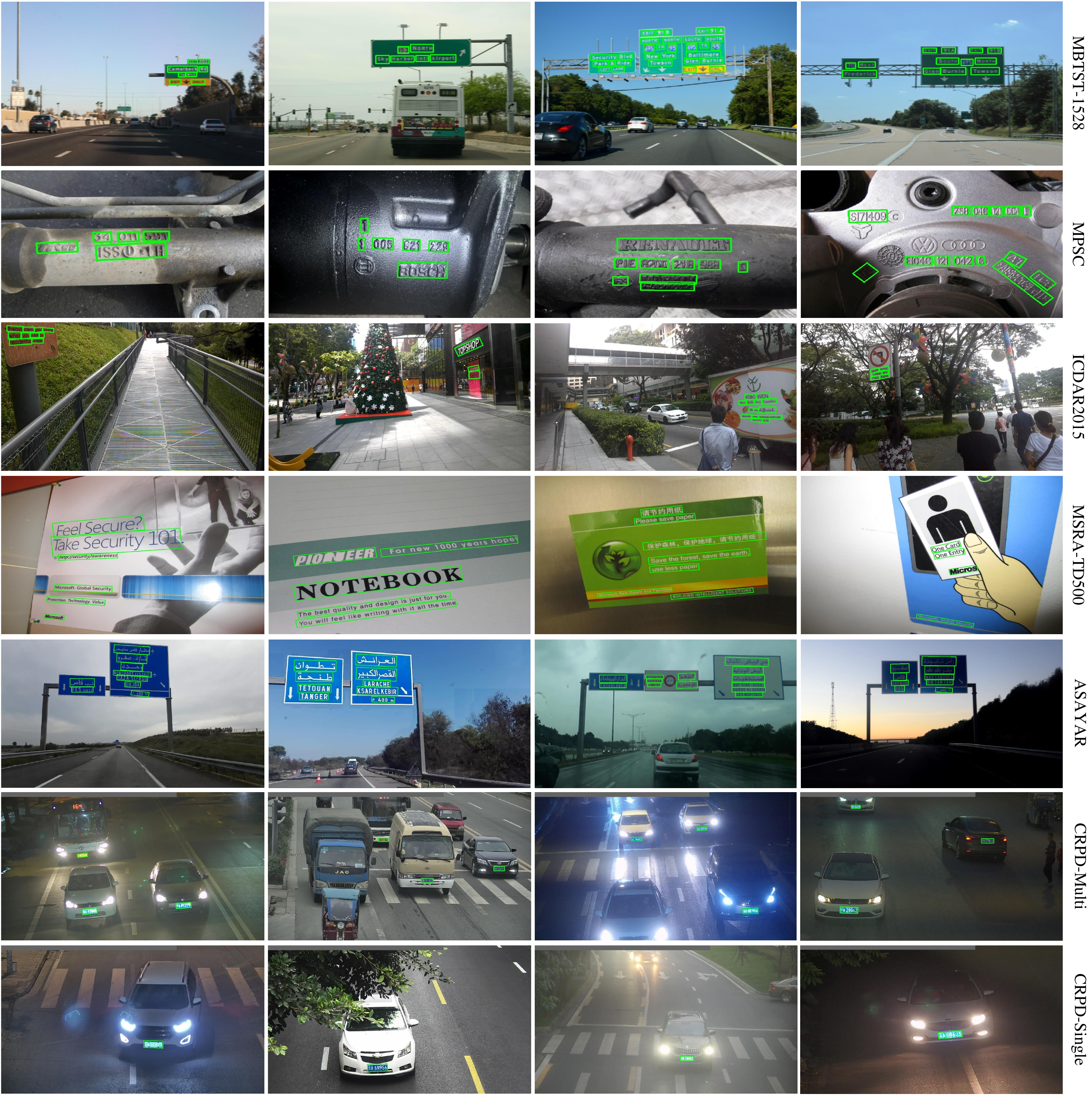}
	\caption{The visualization of detection results. From top to bottom, images come from MBTST-1528, MPSC, ICDAR2015, MSRA-TD500, ASAYAR, CRPD-Multi and CRPD-Single datasets.}
	\label{vis}
\end{figure*}

To verify the superiority of the proposed SMNet, we compare our method with existing SOTA methods on multiple benchmarks including various scenes. In addition, the advantage of SMNet is analyzed based on these comparisons. The prediction of the similar mask and corresponding outputs are visualized in Fig. \ref{sm}. The detection results of SMNet on various datasets are visualized in Fig. \ref{vis}. 
\begin{table}
	\center

	{
		\caption{ Comparison with existing state-of-the-art (SOTA) approaches on the ICDAR2015 dataset. These methods are divided into real-time and non-real-time methods according to the detection speed. ``\textcolor{blue}{\textbf{Blue}}'' and ``\textcolor{red}{\textbf{Red}}'' represent the  best performance of real-time and no-real-time methods, respectively. }
		\setlength{\tabcolsep}{6pt}
		\begin{tabular}{c|c|cccc}

			\hline Methods &Venue  & P & R & F & FPS \\
			\hline 
			DBNet\cite{db} &AAAI'20  &{86.8} &{78.4} &{82.3} &48 \\
			PAN\cite{pan} &ICCV'19  &84.0 &81.9 &82.9 &26.1 \\ 
			PAN++\cite{pan++} &TPAMI'22  &85.9 &80.4 &83.1 &28.2 \\ 
			CMNet\cite{cm} &TIP'22 &86.7 &81.3 &83.9 &34.5 \\
			DBNet++\cite{db++} &TPAMI'23 &90.1 &77.2 &83.1 &44 \\
			
			ZTD \cite{zoom} &TNNLS'23  &87.5 &79.0 &83.0 &48.3\\	
			
			SMNet-Res18&Ours   &{90.3} &{79.3} &\textcolor{blue}{\textbf{84.5}} &48.1 \\ \hline
			DBNet\cite{db} &AAAI'20 &{91.8} &{83.2} &{87.3} &{12}\\ 
			AAM \cite{aam} &TITS'21 &89.8 &85.0 &87.3 &- \\
			FCENet\cite{fcenet} &CVPR'21 &90.1 &82.6 &{86.2} &-\\ 	
			SLN\cite{sln} &TCSVT'21 &88 &83 &85 &11.2 \\
			LEMNet\cite{lem} &TMM'22  &88.3 &85.9 &87.1 &-\\
			ASTD\cite{astd} &TMM'22  &88.8 &82.6 &85.6 &-\\
			EMA\cite{ema} &TIP'22 &89.4 &82.4 &85.8 &21.6 \\
			TextDCT\cite{textdct} &TMM'23   &88.9 &84.8 &{86.8} &7.5\\
			
			DBNet++\cite{db++} &TPAMI'23 &90.9 &{83.9} &{87.3} &{10} \\ 	
			TextPM\cite{textpm} &TPAMI'23 &89.9 &84.9 &87.4 &- \\

			RP-Text \cite{rp} &TMM'23&89.6 &82.4 &85.9 &-\\	
			
			FS \cite{fs} &TIP'23 &89.8 &82.7 &86.4 &12.1 \\
			LeafText\cite{leaftext} &TMM'23 &88.9 &82.3 &86.1 &-\\
			KPN \cite{kpn} &TNNLS'23 &88.3 &84.8 &86.5 &6.3 \\

			SMNet-Res50  &Ours  &{89.7} &85.5 &\textcolor{red}{\textbf{87.6}} &8.9\\

			\hline
		\end{tabular}
		\label{ic15}
	}
	
\end{table}

\begin{table}[t]
	\center	{	
			\caption{ Comparison with existing methods on the ICDAR2013 dataset based on the DetEval standard.  }
			
			\begin{tabular}{c|ccc}
				
				\hline Methods   & Precision & Recall & F-measure  \\ \hline
				CTPN\cite{ctpn}  &93.0 &83.0 &87.7 \\			
				SegLink\cite{seglink}  &87.7 &83.0 &85.3 \\
				PixelLink\cite{pixellink}  &88.6 &87.5 &88.1  \\
				TextBoxes\cite{textbox}   &89 &83 &86  \\
				TextBoxes++\cite{textbox++}   &92 &86 &89  \\
				Boundary\cite{boundary}   &89.3 &85.2 &87.2  \\
				SSTD\cite{sstd} &89 &86 &88  \\
				ASTD\cite{astd} &89.8 &85.1 &87.4\\
				\hline
				SMNet-ResNet18  &{95.5} &{84.6} &89.7  \\
				\hline
			\end{tabular}
			\label{ic13}
			
	}
\end{table}

\begin{table}
	\center

	{
		\caption{ Comparison with existing advanced approaches on the MPSC. The best performance are labeled in \textbf{bold}. $*$ represents the detection results are collected from \cite{rfn}.}
		\begin{tabular}{ccccc}

			\hline Methods  & P & R & F & FPS \\
			\hline 
			
			$ \text{EAST}^*$\cite{east}  & 83.6 & 73.5 & 78.2 & -\\
			$ \text{MASK R-CNN}^*$\cite{maskrcnn}  & 85.3 & 73.5 & 82.2 &- \\
			$ \text{RRPN}^*$\cite{rrpn}  & 82.0 & 78.9 & 80.4  &- \\
			$ \text{PSENet}^*$\cite{pse}  & 85.4 & 78.4 & 81.8  &- \\
			$ \text{PAN}^*$\cite{pan}  & 87.1 & 81.6 & 84.2  &- \\
			$ \text{BDN}^*$\cite{bdn}  & 86.6 &77.5 & 81.8  &- \\
			$ \text{ContourNet}^*$\cite{contournet}  & 87.8 &81.0 & 84.3  &- \\
			$ \text{RRPN++}^*$\cite{rrpn++}  & 86.7 & 83.9 & 85.3  &- \\
			$ \text{FCENet}^*$\cite{fcenet}  & 87.1 & 81.6 & 84.3  &- \\
			RFN-ResNet50\cite{rfn}    &89.3 &83.3 &86.2 &-\\ \hline	
			SMNet-ResNet18     & {88.6} & {82.9} &85.7 & {\textbf{49.4}} \\
			SMNet-ResNet50    & 88.9 & {84.4} & \textbf{86.6} &19.8 \\
			
			\hline
		\end{tabular}
		\label{mpsc}
	}
	
\end{table}
\subsubsection{Evalution on MSRA-TD500} 
We conduct experiments on MSRA-TD500 to prove the performance of the proposed SMNet for detecting multi-oriented line-level texts. We adopt ResNet18 and ResNet50 as the backbone to compare with existing state-of-the-art (SOTA) methods. These methods are classified as real-time and non-real-time according to the detection speed. During the test stage, the short side of the input image is resized to 736. As shown in Table \ref{td500}, the proposed SMNet achieves the best performance and fastest detection speed among many real-time methods. It surpasses the existing SOTA methods RSMTD \cite{rmstd} and DBNet++ \cite{db++} 1.4$\%$ and 2.6$\%$, respectively. When adopting ResNet-50 as the backbone, it achieves 91.0$\%$, 86.8$\%$, and 88.8$\%$ in terms of precision, recall, and F-measure, respectively. It outperforms the advanced methods LOAD \cite{load} and DBNet++ \cite{db++} 1.4$\%$ and 1.6$\%$, respectively. The above experiments further prove the superiority of the proposed SMNet. In addition, we compare the traditional methods in Table \ref{td500_trad}, which demonstrates that the proposed method is superior to the existing traditional methods \cite{Kang_2014_CVPR}, \cite{texstar}, \cite{yao2014unified}, \cite{momv} significantly both in performance and inference speed.
\subsubsection{Evalution on ICDAR2015 Dataset} 
The image is mainly collected from the supermarket, which enjoys a complex background. We conduct experiments on this dataset to verify the robustness of the proposed SM-Net. Similar to MSRA-TD500,  existing SOTA methods are divided into real-time and non-real-time. We follow the DBNet to keep the aspect ratio and resize the short side of the input image to 736 and 1152 when adopting ResNet18 and ResNet50 as the backbone, respectively. Compared with the real-time methods, our method achieves 84.5$\%$ in F-measure, which is the best performance. It outperforms the existing advanced methods ZTD \cite{zoom} and DBNet++ \cite{db++} 1.5$\%$ and 1.4$\%$, respectively. Compared to ResNet18, ResNet50 brings 3.1$\%$ improvements. It surpasses the state-of-the-art methods KPN \cite{kpn} and LeafText \cite{leaftext} 1.1$\%$ and 1.5$\%$, respectively. The above experiments demonstrate that the SM-Net cognizes scene text from complicated backgrounds effectively.

\subsubsection{Evalution on ICDAR2013 Dataset} 
Most of the text in this dataset is horizontal. As shown in Table \ref{ic13}, the proposed method, even with the lightweight ResNet18 backbone, surpasses existing methods. It outperforms SSTD \cite{sstd} by 1.7$\%$ and ASTD \cite{astd} by 2.3$\%$ in F-measure, demonstrating that SM-Net effectively detects horizontal text.
\begin{table}
	\center

	{
		\caption{ Comparison with existing advanced approaches on the CRPD-single. The best performance are labeled in \textbf{bold}.}
		\begin{tabular}{ccccc}

			\hline Methods & P & R & F & FPS \\
			\hline

			SSD512+CRNN \cite{ssd} &28.7 &98.9 &44.4 &71 \\
			YOLOv3+CRNN\cite{yolov3} &59.4 &73.7 &65.8 &18 \\
			YOLOv4+CRNN\cite{yolov4} &68.4 &87.3 &76.7 &40\\
			SYOLOv4+CRNN\cite{syolov4} &72.4 &90.1 &80.3 &35 \\
			Faster-RCNN+CRNN\cite{faster} &71.7 &81.4 &76.3 &20 \\
			STELA+CRNN\cite{stela}&73.3 &83.1 &77.9 &36 \\
			CRPD\cite{crpd} &83.6 &96.3 &89.5 &35 \\
			HFENet\cite{hfe} &97.1 &99.1 &98.1 &35.1 \\ \hline
			SMNet-ResNet18& 98.3 & {98.4} & \textbf{98.4} &\textbf{46.3} \\
			
			\hline
		\end{tabular}
		\label{crpds}
	}
	
\end{table}
\begin{table}
	\center
	\small
	{
		\caption{Comparison with existing advanced approaches on the CRPD-multi. The best performance are labeled in \textbf{bold}.}
		\begin{tabular}{cccccc}

			\hline Methods & P & R & F & FPS \\
			\hline

			SSD512+CRNN\cite{ssd} &21.1 &93.5 &34.5 &63 \\
			YOLOv3+CRNN\cite{yolov3} &61.3 &66.2 &63.6 &17 \\
			YOLOv4+CRNN\cite{yolov4} &36.8 &88.9 &52.0 &39\\
			SYOLOv4+CRNN\cite{syolov4} &75.2 &91.5 &82.5 &35 \\
			Faster-RCNN+CRNN\cite{faster} &75.2 &69.3 &76.3 &17 \\
			STELA+CRNN\cite{stela} &82.8 &77.6 &80.1 &33 \\
			CRPD\cite{crpd} &85.0 &90.8 &87.7 &26 \\
			HFENet\cite{hfe} &95.0 &93.2 &94.1 &35.6 \\ \hline
			SMNet-ResNet18 &95.0  & {93.9} & \textbf{94.4} & {\textbf{43.2}} \\

			\hline
		\end{tabular}
		\label{crpdm}
	}
	
\end{table}

\begin{table}[!t]
	\center

	{
		\caption{ Comparison with existing advanced approaches on the ASAYAR-TXT. The best performance are labeled in \textbf{bold}.}
		\begin{tabular}{ccccc}

			\hline Methods  & Precision & Recall & F-measure & FPS \\
			\hline

			Texboxes++ \cite{asayar} &66 &52 &58 &-\\
			CTPN\cite{asayar}    &80&95 &86 &-\\ 	
			CTPN+Baseline\cite{asayar}    &83&97 &89 &-\\ 
			HFENet\cite{hfe}	&96.3 &97.2 &96.8 &- \\ \hline
			SMNet    & 98.1 & {97.0} & \textbf{97.6} &\textbf{40.4}\\
			
			\hline
		\end{tabular}
		\label{asayar}
	}
	
\end{table}
\begin{table}[t]
	\center

	{
		\caption{Comparison with existing advanced approaches on the MBTST-1528. The best performances are labeled in \textbf{bold}.}
		\begin{tabular}{cccccc}

			\hline Methods &Backbone & P & R & F & FPS \\
			\hline 
			PSE\cite{pse} &ResNet50 &92.4 &76.3 &83.6 &17.2\\
			PAN \cite{pan} &ResNet18 &92.8&79.2 &85.5 &37.8\\
			CT\cite{ct} &ResNet18 &95.2 & 79.7 &86.8 &38.7 \\
			DBNet\cite{db} &ResNet18 &{95.0} &{82.9} &88.5 &66.1\\
			PAN++ \cite{pan++} &ResNet18 &95.7&78.5 &86.2 &21.2\\
			DBNet++\cite{db++} &ResNet18 &{93.6} &{83.5} &88.3 &58.7\\

			\hline
			SMNet   &ResNet18& 96.7 & {84.2} & {\textbf{90.0}} &\textbf{69.8} \\
			
			\hline
		\end{tabular}
		\label{mbtsc}
	}
	
\end{table}

\subsubsection{Evalution on MPSC Dataset }
MPSC mainly contains industrial scene images, which enjoy low contrast and corroded surfaces. This poses a challenge in accurately detecting text. This increases the challenge of detecting industrial texts accurately. We resize the short side of the input images to 800.  As shown in Table \ref{mpsc}, existing SOTA methods RRPN++ \cite{rrpn++}, FCENet \cite{fcenet}, and RFN \cite{rfn} achieve 85.3$\%$, 84.3$\%$, and 86.2$\%$ in terms of F-measure. Benefiting from that the proposed  FCM corrects false prediction at the feature level, our method surpasses them by 1.3$\%$, 2.3$\%$, and 0.4$\%$ in F-measure.  Even if the proposed SMNet adopts a lightweight backbone ResNet-18, it is still superior to most existing methods. Meanwhile, the detection speed is important for industrial applications, and the RFN has a complex framework and a low efficiency that cannot be applied well in real production. The proposed SMNet enjoys a simple network and post-processing, which has a competitive detection speed.

\subsubsection{Evalution on CRPD} 

To prove the generality of the SMNet, we conduct the experiment on the CRPD dataset. We keep the aspect ratio and resize the short size of the input image to 736 on the CRPD dataset. For a fair comparison, we follow the HFENet to evaluate this benchmark and use the SynthText dataset to pre-train. As listed in Table \ref{crpdm} and Table \ref{crpds},  on the CRPD-Single dataset, the proposed SMNet achieves 98.3$\%$, 98.4$\%$, and 98.8$\%$ in precision, recall, and F-measure, respectively. The proposed SMNet surpasses the CRPD \cite{crpd} that establishes the CRPD dataset by 14.7$\%$, 2.1$\%$, and 6.7$\%$ in terms of precision, recall, and F-measure, respectively. Compared with the recent SOTA methods HFENet \cite{hfe}, our method brings 0.3$\%$ gains in F-measure. In addition, our method is superior to it at detection speed. For the CRPD-multi dataset, our method achieves 94.4$\%$ in F-measure, which outperforms existing SOTA methods CRPD and HFENet 6.7$\%$ and 0.3$\%$, respectively. Meanwhile, the proposed method maintains a higher detection efficiency than theirs. The above experiments verify that the proposed method detects  Chinese license plates effectively.

\subsubsection{Evalution on ASAYAR-TXT} 
For images on the ASAYAR-TXT dataset, we follow the HFENet \cite{hfe} to resize the short side of the image to 736 and keep the aspect ratio. We follow the HFENet to utilize the SynthText dataset to pre-train. As shown in Table \ref{asayar}, the proposed SMNet achieves 98.1$\%$, 97.0$\%$, and 97.6$\%$ in terms of precision, recall, and F-measure, which is superior to existing methods. Specifically, our method surpasses CTPN \cite{asayar} and CTPN+baseline \cite{asayar} by 11.6$\%$ and 8.6$\%$ in F-measure. In addition, compared to the recent SOTA methods HFENet, the proposed SMNet outperforms 0.8$\%$ in F-measure while maintaining a competitive detection speed. The above experiments prove that the proposed method achieves SOTA performance on the ASAYAR-TEXT dataset.

\subsubsection{Evalution on MBTST} 
Unlike other traffic text datasets, the collected MBSTC has added the interference of motion blur, which is more challenging than other datasets. All the short side of the input image is resized to 640 while maintaining the aspect ratio. As shown in Table \ref{mbtsc}, the proposed SM-Net achieves  96.7$\%$, 84.2$\%$, and 90.0$\%$ in terms of precision, recall, and F-measure, respectively. Compared to existing SOTA methods DBNet++ and PAN++, our method brings 1.7$\%$ and 3.8$\%$ improvements in F-measure. In addition, the proposed SMNet achieves the fastest detection speed. The experiments demonstrate the proposed method enjoys great anti-interference capability.
\subsubsection{Evalution on TGPD}
Since its training images and corresponding ground truth are not available, we train on the proposed MBTST dataset. The testing annotation is sometimes confusing. As shown in Fig. \ref{tgpt}, the error annotations are labeled in red, and some missing annotations are labeled in orange. This further proves the necessity of the proposed dataset. The detection results are listed in Table \ref{tgpd}. Note that,  \cite{zhu2017cascaded} trains on the TTSDCE dataset, which is also not available. We keep the aspect ratio and resize the short side of the input image to 736. The proposed method adopts ResNet18 as the backbone, which is still superior to HFENet and AAM using ResNet50.

\begin{table}[t]
	\center
	\setlength{\tabcolsep}{1.5mm}
	
	{
		\caption{Comparison with existing advanced approaches on the TGPD. $\dagger$ means to adopt the correct ground truth. $*$ represents the another network framework. }
		\begin{tabular}{ccccccc}

			\hline Methods &Backbone& Data & P & R & F & FPS \\
			\hline 
			Rong \emph{et al.}\cite{tgpd} &VGG16 &TGPD &{73} &{64} &68 &6.25\\
			Zhu \emph{et al.}\cite{zhu2017cascaded} &VGG16 &TTSDCE &{90} &{87} &88 &6.67\\
			AAM \cite{aam} &ResNet50 &MLT&70.3 &78.9 &74.4 &- \\
			$\text{AAM}^{\dagger}$\cite{aam} &ResNet50 &MLT &66.9 &85.6 &75.1 &- \\
			HFENet\cite{hfe} &ResNet50 &MLT &72.6 &82.3 &77.1 &12.2\\
			$\text{HFENet}^*$\cite{hfe} &ResNet50 &MLT &72.1 &81.8 &76.7 &12.2\\

			\hline
			SMNet   &ResNet18&MBTST &80.1 & {79.6} & 79.8 &32 \\
			
			\hline
		\end{tabular}
		\label{tgpd}
	}
	
\end{table}

\begin{table}[t]

\center	{
	{
		\caption{Cross-datasets experiments. ``Train'' and ``Test'' represent the train and test datasets. }
		\begin{tabular}{ccccccccc}

			\hline 
		Methods	&	Train	 &Test &Precision &Recall &F-measure    \\ \hline
		\multirow{2}{*}{DBNet++}  &ICDAR2015 &MBTST  &89.9 &62.0 &73.4 \\ 
	  &ICDAR2015 &MPSC   &73.3 &45.5 &56.2\\ 
	\multirow{2}{*}{SMNet}  &ICDAR2015 &MBTST &91.6 &70.1 &79.4  \\ 
		  &ICDAR2015 &MPSC  &73.3 &51.4 &60.5 \\ 

			\hline 
		\end{tabular}
		\label{cross}
	}
}
\end{table}

\subsection{Generalizability Analaysis}
Inspired by SST \cite{SST} and SSH \cite{wang2016measure}, we analyze the generalizability of the proposed module from the properties of the population, the condition of the sample, and the method of estimation. The population comprises images collected from natural, industrial, and traffic scenes, which contain complex backgrounds such as machinery, trees, grass, roads, and vehicles. Additionally, the used datasets have a sufficient number of samples for different scenes to validate the model's performance. Specifically, the natural, industrial, and traffic scenes have 3, 1, and 3 datasets and 1023, 639, and 1257 images, respectively.
For the method of estimation, we perform a cross-dataset experiment. The model is trained on the natural scene dataset ICDAR2015 and tested on the industrial scene dataset MPSC and the traffic scene dataset MBTST. As shown in Table \ref{cross}, the SMNet achieves 79.4$\%$ and 60.5$\%$ on MPSC and MBTST, respectively, significantly outperforming the existing SOTA method DBNet++ \cite{db++}. This experiment further verifies the generalizability of the proposed method.

\section{Conclusion}
\label{conclusion}
In this paper, we establish a challenging traffic motion blur text dataset and propose an innovative and efficient multi-scene text detector. Specifically, we consider the motion blur in the real world and add it to the proposed dataset. The proposed method consists of a new text presentation similar mask and a feature correction module. Unlike the previous text representation methods, the former keeps geometry contour information as much as possible. Meanwhile, it enjoys more efficient and robust post-processing, which accelerates the inference procedure. The latter helps the model correct the false positive predictions at the feature level, which promotes the model recognizing the similar mask from the complicated background. Particularly, the proposed SMNet achieves 88.8$\%$, 87.6$\%$, 86.6$\%$, 98.4$\%$, 94.4$\%$, and 97.6$\%$ in F-measure for MSRA-TD500, ICDAR2015, MPSC, CRPD-Single, CRPD-Multi, and ASAYAR-TXT datasets, respectively. Our method can be applied to scene text detection, traffic plane text detection, and license plate detection, which provide assistance for other tasks (such as autonomous driving, scene analysis, and intelligent transportation).

\ifCLASSOPTIONcaptionsoff
\newpage
\fi
\linespread{1}
\bibliographystyle{IEEETran}
\bibliography{IEEEabrv}

\begin{thebibliography}{10}
\providecommand{\url}[1]{#1}
\csname url@samestyle\endcsname
\providecommand{\newblock}{\relax}
\providecommand{\bibinfo}[2]{#2}
\providecommand{\BIBentrySTDinterwordspacing}{\spaceskip=0pt\relax}
\providecommand{\BIBentryALTinterwordstretchfactor}{4}
\providecommand{\BIBentryALTinterwordspacing}{\spaceskip=\fontdimen2\font plus
\BIBentryALTinterwordstretchfactor\fontdimen3\font minus
  \fontdimen4\font\relax}
\providecommand{\BIBforeignlanguage}[2]{{%
\expandafter\ifx\csname l@#1\endcsname\relax
\typeout{** WARNING: IEEEtran.bst: No hyphenation pattern has been}%
\typeout{** loaded for the language `#1'. Using the pattern for}%
\typeout{** the default language instead.}%
\else
\language=\csname l@#1\endcsname
\fi
#2}}
\providecommand{\BIBdecl}{\relax}
\BIBdecl

\bibitem{traffic}
C.~Yang, K.~Zhuang, M.~Chen, H.~Ma, X.~Han, T.~Han, C.~Guo, H.~Han, B.~Zhao,
  and Q.~Wang, ``Traffic sign interpretation via natural language
  description,'' \emph{IEEE Transactions on Intelligent Transportation
  Systems}, pp. 1--15, 2024.

\bibitem{rfn}
T.~Guan, C.~Gu, C.~Lu, J.~Tu, Q.~Feng, K.~Wu, and X.~Guan, ``Industrial scene
  text detection with refined feature-attentive network,'' \emph{IEEE
  Transactions on Circuits and Systems for Video Technology}, vol.~32, no.~9,
  pp. 6073--6085, 2022.

\bibitem{tgpd}
X.~Rong, C.~Yi, and Y.~Tian, ``Recognizing text-based traffic guide panels with
  cascaded localization network,'' in \emph{Computer Vision--ECCV 2016
  Workshops: Amsterdam, The Netherlands, October 8-10 and 15-16, 2016,
  Proceedings, Part I 14}.\hskip 1em plus 0.5em minus 0.4em\relax Springer,
  2016, pp. 109--121.

\bibitem{zhu2017cascaded}
Y.~Zhu, M.~Liao, M.~Yang, and W.~Liu, ``Cascaded segmentation-detection
  networks for text-based traffic sign detection,'' \emph{IEEE transactions on
  intelligent transportation systems}, vol.~19, no.~1, pp. 209--219, 2017.

\bibitem{pse}
W.~Wang, E.~Xie, X.~Li, W.~Hou, T.~Lu, G.~Yu, and S.~Shao, ``Shape robust text
  detection with progressive scale expansion network,'' in \emph{Proceedings of
  the IEEE Conference on Computer Vision and Pattern Recognition}, 2019, pp.
  9336--9345.

\bibitem{pan}
W.~Wang, E.~Xie, X.~Song, Y.~Zang, W.~Wang, T.~Lu, G.~Yu, and C.~Shen,
  ``Efficient and accurate arbitrary-shaped text detection with pixel
  aggregation network,'' in \emph{Proceedings of the IEEE International
  Conference on Computer Vision}, 2019, pp. 8440--8449.

\bibitem{db}
M.~Liao, Z.~Wan, C.~Yao, K.~Chen, and X.~Bai, ``Real-time scene text detection
  with differentiable binarization,'' in \emph{Proceedings of the AAAI
  Conference on Artificial Intelligence}, vol.~34, no.~07, 2020, pp.
  11\,474--11\,481.

\bibitem{vatti1992generic}
B.~R. Vatti, ``A generic solution to polygon clipping,'' \emph{Communications
  of the ACM}, vol.~35, no.~7, pp. 56--63, 1992.

\bibitem{ad}
Y.~Qu, H.~Xie, S.~Fang, Y.~Wang, and Y.~Zhang, ``Adnet: Rethinking the shrunk
  polygon-based approach in scene text detection,'' \emph{IEEE Transactions on
  Multimedia}, pp. 1--14, 2022.

\bibitem{rmstd}
C.~Yang, M.~Chen, Y.~Yuan, and Q.~Wang, ``Reinforcement shrink-mask for text
  detection,'' \emph{IEEE Transactions on Multimedia}, vol.~25, pp. 6458--6470,
  2023.

\bibitem{faster}
S.~Ren, K.~He, R.~Girshick, and J.~Sun, ``Faster r-cnn: Towards real-time
  object detection with region proposal networks,'' \emph{Advances in neural
  information processing systems}, vol.~28, pp. 91--99, 2015.

\bibitem{rrpn}
J.~Ma, W.~Shao, H.~Ye, L.~Wang, H.~Wang, Y.~Zheng, and X.~Xue,
  ``Arbitrary-oriented scene text detection via rotation proposals,''
  \emph{IEEE Trans. Multimedia}, vol.~20, no.~11, pp. 3111--3122, 2018.

\bibitem{east}
X.~Zhou, C.~Yao, H.~Wen, Y.~Wang, S.~Zhou, W.~He, and J.~Liang, ``East: an
  efficient and accurate scene text detector,'' in \emph{Proceedings of the
  IEEE conference on Computer Vision and Pattern Recognition}, 2017, pp.
  5551--5560.

\bibitem{textbox}
M.~Liao, B.~Shi, X.~Bai, X.~Wang, and W.~Liu, ``Textboxes: A fast text detector
  with a single deep neural network,'' in \emph{Thirty-first AAAI conference on
  artificial intelligence}, 2017.

\bibitem{ssd}
\BIBentryALTinterwordspacing
W.~Liu, D.~Anguelov, D.~Erhan, C.~Szegedy, S.~E. Reed, C.~Fu, and A.~C. Berg,
  ``{SSD:} single shot multibox detector,'' in \emph{Computer Vision - {ECCV}
  2016 - 14th European Conference, Amsterdam, The Netherlands, October 11-14,
  2016, Proceedings, Part {I}}, ser. Lecture Notes in Computer Science, vol.
  9905.\hskip 1em plus 0.5em minus 0.4em\relax Springer, 2016, pp. 21--37.
  [Online]. Available: \url{https://doi.org/10.1007/978-3-319-46448-0\_2}
\BIBentrySTDinterwordspacing

\bibitem{textbox++}
M.~Liao, B.~Shi, and X.~Bai, ``Textboxes++: A single-shot oriented scene text
  detector,'' \emph{IEEE transactions on image processing}, vol.~27, no.~8, pp.
  3676--3690, 2018.

\bibitem{dmpn}
Y.~Liu and L.~Jin, ``Deep matching prior network: Toward tighter multi-oriented
  text detection,'' in \emph{Proceedings of the IEEE conference on computer
  vision and pattern recognition}, 2017, pp. 1962--1969.

\bibitem{rrd}
M.~Liao, Z.~Zhu, B.~Shi, G.-s. Xia, and X.~Bai, ``Rotation-sensitive regression
  for oriented scene text detection,'' in \emph{Proceedings of the IEEE
  Conference on Computer Vision and Pattern Recognition (CVPR)}, June 2018.

\bibitem{abcnet}
Y.~Liu, H.~Chen, C.~Shen, T.~He, L.~Jin, and L.~Wang, ``Abcnet: Real-time scene
  text spotting with adaptive bezier-curve network,'' in \emph{proceedings of
  the IEEE conference on computer vision and pattern recognition}, 2020, pp.
  9809--9818.

\bibitem{fcenet}
Y.~Zhu, J.~Chen, L.~Liang, Z.~Kuang, L.~Jin, and W.~Zhang, ``Fourier contour
  embedding for arbitrary-shaped text detection,'' in \emph{Proceedings of the
  IEEE Conference on Computer Vision and Pattern Recognition}, 2021, pp.
  3123--3131.

\bibitem{textsnake}
S.~Long, J.~Ruan, W.~Zhang, X.~He, W.~Wu, and C.~Yao, ``Textsnake: A flexible
  representation for detecting text of arbitrary shapes,'' in \emph{Proceedings
  of the European conference on computer vision}, 2018, pp. 20--36.

\bibitem{textdragon}
W.~Feng, W.~He, F.~Yin, X.-Y. Zhang, and C.-L. Liu, ``Textdragon: An end-to-end
  framework for arbitrary shaped text spotting,'' in \emph{Proceedings of the
  IEEE/CVF international conference on computer vision}, 2019, pp. 9076--9085.

\bibitem{drrg}
S.-X. Zhang, X.~Zhu, J.-B. Hou, C.~Liu, C.~Yang, H.~Wang, and X.-C. Yin, ``Deep
  relational reasoning graph network for arbitrary shape text detection,'' in
  \emph{Proceedings of the IEEE/CVF Conference on Computer Vision and Pattern
  Recognition (CVPR)}, June 2020.

\bibitem{seglink}
B.~Shi, X.~Bai, and S.~Belongie, ``Detecting oriented text in natural images by
  linking segments,'' in \emph{Proceedings of the IEEE conference on computer
  vision and pattern recognition}, 2017, pp. 2550--2558.

\bibitem{pixellink}
D.~Deng, H.~Liu, X.~Li, and D.~Cai, ``Pixellink: Detecting scene text via
  instance segmentation,'' in \emph{Proceedings of the AAAI Conference on
  Artificial Intelligence}, vol.~32, no.~1, 2018.

\bibitem{spot}
\BIBentryALTinterwordspacing
X.~Han, J.~Gao, C.~Yang, Y.~Yuan, and Q.~Wang, ``Spotlight text detector:
  Spotlight on candidate regions like a camera,'' 2024. [Online]. Available:
  \url{https://arxiv.org/abs/2409.16820}
\BIBentrySTDinterwordspacing

\bibitem{fepe}
\BIBentryALTinterwordspacing
------, ``Focus entirety and perceive environment for arbitrary-shaped text
  detection,'' 2024. [Online]. Available:
  \url{https://arxiv.org/abs/2409.16827}
\BIBentrySTDinterwordspacing

\bibitem{leaftext}
C.~Yang, M.~Chen, Y.~Yuan, and Q.~Wang, ``Text growing on leaf,'' \emph{IEEE
  Transactions on Multimedia}, vol.~25, pp. 9029--9043, 2023.

\bibitem{tkc}
X.~Han, J.~Gao, Y.~Yuan, and Q.~Wang, ``Text kernel calculation for arbitrary
  shape text detection,'' \emph{The Visual Computer}, vol.~40, no.~4, pp.
  2641--2654, 2024.

\bibitem{aam}
J.-B. Hou, X.~Zhu, C.~Liu, C.~Yang, L.-H. Wu, H.~Wang, and X.-C. Yin,
  ``Detecting text in scene and traffic guide panels with attention anchor
  mechanism,'' \emph{IEEE Transactions on Intelligent Transportation Systems},
  vol.~22, no.~11, pp. 6890--6899, 2021.

\bibitem{hfe}
M.~Liang, X.~Zhu, H.~Zhou, J.~Qin, and X.-C. Yin, ``Hfenet: Hybrid feature
  enhancement network for detecting texts in scenes and traffic panels,''
  \emph{IEEE Transactions on Intelligent Transportation Systems}, vol.~24,
  no.~12, pp. 14\,200--14\,212, 2023.

\bibitem{karatzas2015icdar}
D.~Karatzas, L.~Gomez-Bigorda, A.~Nicolaou, S.~Ghosh, A.~Bagdanov, M.~Iwamura,
  J.~Matas, L.~Neumann, V.~R. Chandrasekhar, S.~Lu \emph{et~al.}, ``Icdar 2015
  competition on robust reading,'' in \emph{2015 13th International Conference
  on Document Analysis and Recognition}.\hskip 1em plus 0.5em minus 0.4em\relax
  IEEE, 2015, pp. 1156--1160.

\bibitem{yao2012detecting}
C.~Yao, X.~Bai, W.~Liu, Y.~Ma, and Z.~Tu, ``Detecting texts of arbitrary
  orientations in natural images,'' in \emph{2012 IEEE conference on computer
  vision and pattern recognition}.\hskip 1em plus 0.5em minus 0.4em\relax IEEE,
  2012, pp. 1083--1090.

\bibitem{yao2014unified}
C.~Yao, X.~Bai, and W.~Liu, ``A unified framework for multioriented text
  detection and recognition,'' \emph{IEEE Transactions on Image Processing},
  vol.~23, no.~11, pp. 4737--4749, 2014.

\bibitem{asayar}
M.~Akallouch, K.~S. Boujemaa, A.~Bouhoute, K.~Fardousse, and I.~Berrada,
  ``Asayar: A dataset for arabic-latin scene text localization in highway
  traffic panels,'' \emph{IEEE Transactions on Intelligent Transportation
  Systems}, vol.~23, no.~4, pp. 3026--3036, 2022.

\bibitem{crpd}
Y.~Gong, L.~Deng, S.~Tao, X.~Lu, P.~Wu, Z.~Xie, Z.~Ma, and M.~Xie, ``Unified
  chinese license plate detection and recognition with high efficiency,''
  \emph{Journal of Visual Communication and Image Representation}, vol.~86, p.
  103541, 2022.

\bibitem{fpn}
T.~Lin, P.~Doll{\'a}r, R.~Girshick, K.~He, B.~Hariharan, and S.~Belongie,
  ``Feature pyramid networks for object detection,'' in \emph{Proceedings of
  the IEEE conference on computer vision and pattern recognition}, 2017, pp.
  2117--2125.

\bibitem{resnet}
K.~He, X.~Zhang, S.~Ren, and J.~Sun, ``Deep residual learning for image
  recognition,'' in \emph{Proceedings of the IEEE conference on computer vision
  and pattern recognition}, 2016, pp. 770--778.

\bibitem{deformable2}
X.~Zhu, H.~Hu, S.~Lin, and J.~Dai, ``Deformable convnets v2: More deformable,
  better results,'' in \emph{Proceedings of the IEEE conference on computer
  vision and pattern recognition}, 2019, pp. 9308--9316.

\bibitem{gupta2016synthetic}
A.~Gupta, A.~Vedaldi, and A.~Zisserman, ``Synthetic data for text localisation
  in natural images,'' in \emph{Proceedings of the IEEE conference on computer
  vision and pattern recognition}, 2016, pp. 2315--2324.

\bibitem{texstar}
X.-C. Yin, X.~Yin, K.~Huang, and H.-W. Hao, ``Robust text detection in natural
  scene images,'' \emph{IEEE Transactions on Pattern Analysis and Machine
  Intelligence}, vol.~36, no.~5, pp. 970--983, 2014.

\bibitem{ct}
T.~Sheng, J.~Chen, Z.~Lian, and q.~qzz, ``Centripetaltext: An efficient text
  instance representation for scene text detection,'' \emph{Advances in Neural
  Information Processing Systems}, vol.~34, pp. 335--346, 2021.

\bibitem{pan++}
W.~Wang, E.~Xie, X.~Li, X.~Liu, D.~Liang, Z.~Yang, T.~Lu, and C.~Shen, ``Pan++:
  Towards efficient and accurate end-to-end spotting of arbitrarily-shaped
  text,'' \emph{IEEE Trans. Pattern Analysis and Machine Intelligence},
  vol.~44, no.~9, pp. 5349--5367, 2022.

\bibitem{cm}
C.~Yang, M.~Chen, Z.~Xiong, Y.~Yuan, and Q.~Wang, ``Cm-net: Concentric mask
  based arbitrary-shaped text detection,'' \emph{IEEE Transactions on Image
  Processing}, pp. 2864--2877, 2022.

\bibitem{zoom}
C.~Yang, M.~Chen, Y.~Yuan, and Q.~Wang, ``Zoom text detector,'' \emph{IEEE
  Trans. Neural Networks and Learning Systems}, 2023, early Access, doi:
  10.1109/TNNLS.2023.3289327.

\bibitem{db++}
M.~Liao, Z.~Zou, Z.~Wan, C.~Yao, and X.~Bai, ``Real-time scene text detection
  with differentiable binarization and adaptive scale fusion,'' \emph{IEEE
  Transactions on Pattern Analysis and Machine Intelligence}, vol.~45, no.~1,
  pp. 919--931, 2023.

\bibitem{craft}
Y.~Baek, B.~Lee, D.~Han, S.~Yun, and H.~Lee, ``Character region awareness for
  text detection,'' in \emph{Proceedings of the IEEE Conference on Computer
  Vision and Pattern Recognition}, 2019, pp. 9365--9374.

\bibitem{lem}
M.~Xing, H.~Xie, Q.~Tan, S.~Fang, Y.~Wang, Z.~Zha, and Y.~Zhang,
  ``Boundary-aware arbitrary-shaped scene text detector with learnable
  embedding network,'' \emph{IEEE Transactions on Multimedia}, vol.~24, pp.
  3129--3143, 2022.

\bibitem{ema}
M.~Zhao, W.~Feng, F.~Yin, X.-Y. Zhang, and C.-L. Liu, ``Mixed-supervised scene
  text detection with expectation-maximization algorithm,'' \emph{IEEE
  Transactions on Image Processing}, vol.~31, pp. 5513--5528, 2022.

\bibitem{rp}
Q.~Wang, B.~Fu, M.~Li, J.~He, X.~Peng, and Y.~Qiao, ``Region-aware
  arbitrary-shaped text detection with progressive fusion,'' \emph{IEEE
  Transactions on Multimedia}, vol.~25, pp. 4718--4729, 2023.

\bibitem{morph}
C.~Xu, W.~Jia, R.~Wang, X.~Luo, and X.~He, ``Morphtext: Deep morphology
  regularized accurate arbitrary-shape scene text detection,'' \emph{IEEE
  Transactions on Multimedia}, vol.~25, pp. 4199--4212, 2023.

\bibitem{vrrca}
C.~Xu, W.~Jia, T.~Cui, R.~Wang, Y.-f. Zhang, and X.~He, ``Arbitrary-shape scene
  text detection via visual-relational rectification and contour
  approximation,'' \emph{IEEE Transactions on Multimedia}, vol.~25, pp.
  4052--4066, 2023.

\bibitem{fs}
F.~Wang, X.~Xu, Y.~Chen, and X.~Li, ``Fuzzy semantics for arbitrary-shaped
  scene text detection,'' \emph{IEEE Transactions on Image Processing},
  vol.~32, pp. 1--12, 2023.

\bibitem{load}
P.~Cheng, Y.~Zhao, and W.~Wang, ``Detect arbitrary-shaped text via adaptive
  thresholding and localization quality estimation,'' \emph{IEEE Transactions
  on Circuits and Systems for Video Technology}, vol.~33, no.~12, pp.
  7480--7490, 2023.

\bibitem{Kang_2014_CVPR}
L.~Kang, Y.~Li, and D.~Doermann, ``Orientation robust text line detection in
  natural images,'' in \emph{Proceedings of the IEEE Conference on Computer
  Vision and Pattern Recognition (CVPR)}, June 2014.

\bibitem{momv}
X.-C. Yin, W.-Y. Pei, J.~Zhang, and H.-W. Hao, ``Multi-orientation scene text
  detection with adaptive clustering,'' \emph{IEEE Transactions on Pattern
  Analysis and Machine Intelligence}, vol.~37, no.~9, pp. 1930--1937, 2015.

\bibitem{sln}
Y.~Cai, C.~Liu, P.~Cheng, D.~Du, L.~Zhang, W.~Wang, and Q.~Ye, ``Scale-residual
  learning network for scene text detection,'' \emph{IEEE Transactions on
  Circuits and Systems for Video Technology}, vol.~31, no.~7, pp. 2725--2738,
  2021.

\bibitem{astd}
P.~Dai, Y.~Li, H.~Zhang, J.~Li, and X.~Cao, ``Accurate scene text detection via
  scale-aware data augmentation and shape similarity constraint,'' \emph{IEEE
  Transactions on Multimedia}, 2021.

\bibitem{textdct}
Y.~Su, Z.~Shao, Y.~Zhou, F.~Meng, H.~Zhu, B.~Liu, and R.~Yao, ``Textdct:
  Arbitrary-shaped text detection via discrete cosine transform mask,''
  \emph{IEEE Transactions on Multimedia}, pp. 1--14, 2022.

\bibitem{textpm}
S.-X. Zhang, X.~Zhu, L.~Chen, J.-B. Hou, and X.-C. Yin, ``Arbitrary shape text
  detection via segmentation with probability maps,'' \emph{IEEE Transactions
  on Pattern Analysis and Machine Intelligence}, vol.~45, no.~3, pp.
  2736--2750, 2023.

\bibitem{kpn}
S.~Zhang, X.~Zhu, J.~Hou, C.~Yang, and X.~Yin, ``Kernel proposal network for
  arbitrary shape text detection,'' \emph{IEEE Transactions on Neural Networks
  and Learning Systems}, pp. 1--12, 2022.

\bibitem{ctpn}
Z.~Tian, W.~Huang, T.~He, P.~He, and Y.~Qiao, ``Detecting text in natural image
  with connectionist text proposal network,'' in \emph{Proc. ECCV}.\hskip 1em
  plus 0.5em minus 0.4em\relax Springer, 2016, pp. 56--72.

\bibitem{boundary}
H.~Wang, P.~Lu, H.~Zhang, M.~Yang, X.~Bai, Y.~Xu, M.~He, Y.~Wang, and W.~Liu,
  ``All you need is boundary: Toward arbitrary-shaped text spotting,'' in
  \emph{Proceedings of the AAAI conference on artificial intelligence},
  vol.~34, no.~07, 2020, pp. 12\,160--12\,167.

\bibitem{sstd}
P.~He, W.~Huang, T.~He, Q.~Zhu, Y.~Qiao, and X.~Li, ``Single shot text detector
  with regional attention,'' in \emph{Proceedings of the IEEE international
  conference on computer vision}, 2017, pp. 3047--3055.

\bibitem{maskrcnn}
K.~He, G.~Gkioxari, P.~Doll{\'a}r, and R.~Girshick, ``Mask r-cnn,'' in
  \emph{Proceedings of the IEEE international conference on computer vision},
  2017, pp. 2961--2969.

\bibitem{bdn}
Y.~Liu, T.~He, H.~Chen, X.~Wang, C.~Luo, S.~Zhang, C.~Shen, and L.~Jin,
  ``Exploring the capacity of an orderless box discretization network for
  multi-orientation scene text detection,'' \emph{International Journal of
  Computer Vision}, vol. 129, pp. 1972--1992, 2021.

\bibitem{contournet}
Y.~Wang, H.~Xie, Z.-J. Zha, M.~Xing, Z.~Fu, and Y.~Zhang, ``Contournet: Taking
  a further step toward accurate arbitrary-shaped scene text detection,'' in
  \emph{proceedings of the IEEE/CVF conference on computer vision and pattern
  recognition}, 2020, pp. 11\,753--11\,762.

\bibitem{rrpn++}
J.~Ma, ``Rrpn++: Guidance towards more accurate scene text detection,''
  \emph{arXiv preprint arXiv:2009.13118}, 2020.

\bibitem{yolov3}
J.~Redmon and A.~Farhadi, ``Yolov3: An incremental improvement,'' \emph{arXiv
  preprint arXiv:1804.02767}, 2018.

\bibitem{yolov4}
A.~Bochkovskiy, C.-Y. Wang, and H.-Y.~M. Liao, ``Yolov4: Optimal speed and
  accuracy of object detection,'' \emph{arXiv preprint arXiv:2004.10934}, 2020.

\bibitem{syolov4}
C.-Y. Wang, A.~Bochkovskiy, and H.-Y.~M. Liao, ``Scaled-yolov4: Scaling cross
  stage partial network,'' in \emph{Proceedings of the IEEE/cvf conference on
  computer vision and pattern recognition}, 2021, pp. 13\,029--13\,038.

\bibitem{stela}
L.~Deng, Y.~Gong, X.~Lu, Y.~Lin, Z.~Ma, and M.~Xie, ``Stela: A real-time scene
  text detector with learned anchor,'' \emph{IEEE Access}, vol.~7, pp.
  153\,400--153\,407, 2019.

\bibitem{SST}
J.~Wang, B.~Gao, and A.~Stein, ``The spatial statistic trinity: A generic
  framework for spatial sampling and inference,'' \emph{Environmental Modelling
  \& Software}, vol. 134, p. 104835, 2020.

\bibitem{wang2016measure}
J.-F. Wang, T.-L. Zhang, and B.-J. Fu, ``A measure of spatial stratified
  heterogeneity,'' \emph{Ecological indicators}, vol.~67, pp. 250--256, 2016.

\end{thebibliography}

\end{document}